\documentclass[pdflatex,sn-mathphys-num]{sn-jnl}
\usepackage{graphicx}%
\usepackage{multirow}%
\usepackage{amsmath,amssymb,amsfonts}%
\usepackage{amsthm}%
\usepackage{mathrsfs}%
\usepackage[title]{appendix}%
\usepackage{xcolor}%
\usepackage{textcomp}%
\usepackage{manyfoot}%
\usepackage{booktabs}%
\usepackage{algorithm}%
\usepackage{algorithmicx}%
\usepackage{algpseudocode}%
\usepackage{listings}%
\usepackage{lineno}
\usepackage{longtable}

\begin{document}

\title[Article Title]{Noise-shaped Sampling and Annealing for Scalable On-the-edge Training of Binary and Ternary AI Models}

\author[1]{\fnm{Zihao} \sur{Chen}}
\equalcont{These authors contributed equally to this work.}
\author[1]{\fnm{Zhili} \sur{Xiao}}
\equalcont{These authors contributed equally to this work.}
\author[2]{\fnm{Rui-Jie} \sur{Zhu}}
\author[3]{\fnm{Yipeng} \sur{Wang}}
\author[2]{\fnm{Jason} \sur{Eshraghian}}
\author*[1]{\fnm{Shantanu} \sur{Chakrabartty}}\email{shantanu@wustl.edu}

\affil[1]{\orgdiv{Department of Electrical and Systems Engineering}, \orgname{Washington University in St. Louis}, \orgaddress{\street{One Brookings Drive}, \city{St. Louis}, \postcode{63130}, \state{MO}, \country{USA}}}

\affil[2]{\orgdiv{Department of Electrical and Computer Engineering}, \orgname{University of California, Santa Cruz}, \orgaddress{\street{1156 High Street}, \city{Santa Cruz}, \postcode{95064}, \state{CA}, \country{USA}}}

\affil[3]{\orgdiv{Department of Computer Science and Engineering}, \orgname{Washington University in St. Louis}, \orgaddress{\street{One Brookings Drive}, \city{St. Louis}, \postcode{63130}, \state{MO}, \country{USA}}}

\abstract{The training loss of binary and ternary neural networks, expanded as a multilinear polynomial over the discrete weights, is structurally equivalent to the energy of a hierarchical high-order Ising model. We exploit this equivalence to apply Ising solvers to discrete-weight network training, while accounting for a fundamental difference in objectives: an Ising solver seeks the ground state of a fixed energy function, whereas training seeks solutions that generalize to unseen data. Our key contribution is $\Sigma\Delta$-Sampler, an optimizer that reconciles these objectives by embedding a stochastic sampling-and-annealing step as a noisy quantizer inside a first-order $\Sigma\Delta$ noise-shaping loop. The loop pushes noise from weight quantization, high-order frustration, and mini-batch sampling out of the low-frequency adaptation band, so that discrete weight updates follow the smooth descent trajectories embedded in the rugged loss landscape. We validate the framework on a planted quadratic unconstrained binary optimization (QUBO) benchmark with a known minimizer; on CIFAR-10, where a $13$M-parameter ternary model comes within $1\%$ of full-precision accuracy; on ImageNet-1k, where a ternary ResNet-18 comes within $0.4\%$ Top-5 accuracy of a ternary baseline; on class-incremental continual learning, where prior-task accuracy is preserved; and on MatMul-free language models up to $2.7$B parameters, where it outperforms a straight-through-estimator baseline under matched continued pre-training. Compared with standard quantization-aware training, $\Sigma\Delta$-Sampler reduces the training memory footprint by over $50\%$, delivers an order-of-magnitude speed-up, and reduces weight-update memory traffic by four orders of magnitude, making it attractive for on-device training and fine-tuning of large binary and ternary models.}


\maketitle

\section{Introduction}\label{sec1}

\begin{figure}[h!]
\centering
\includegraphics[width=0.99\textwidth]{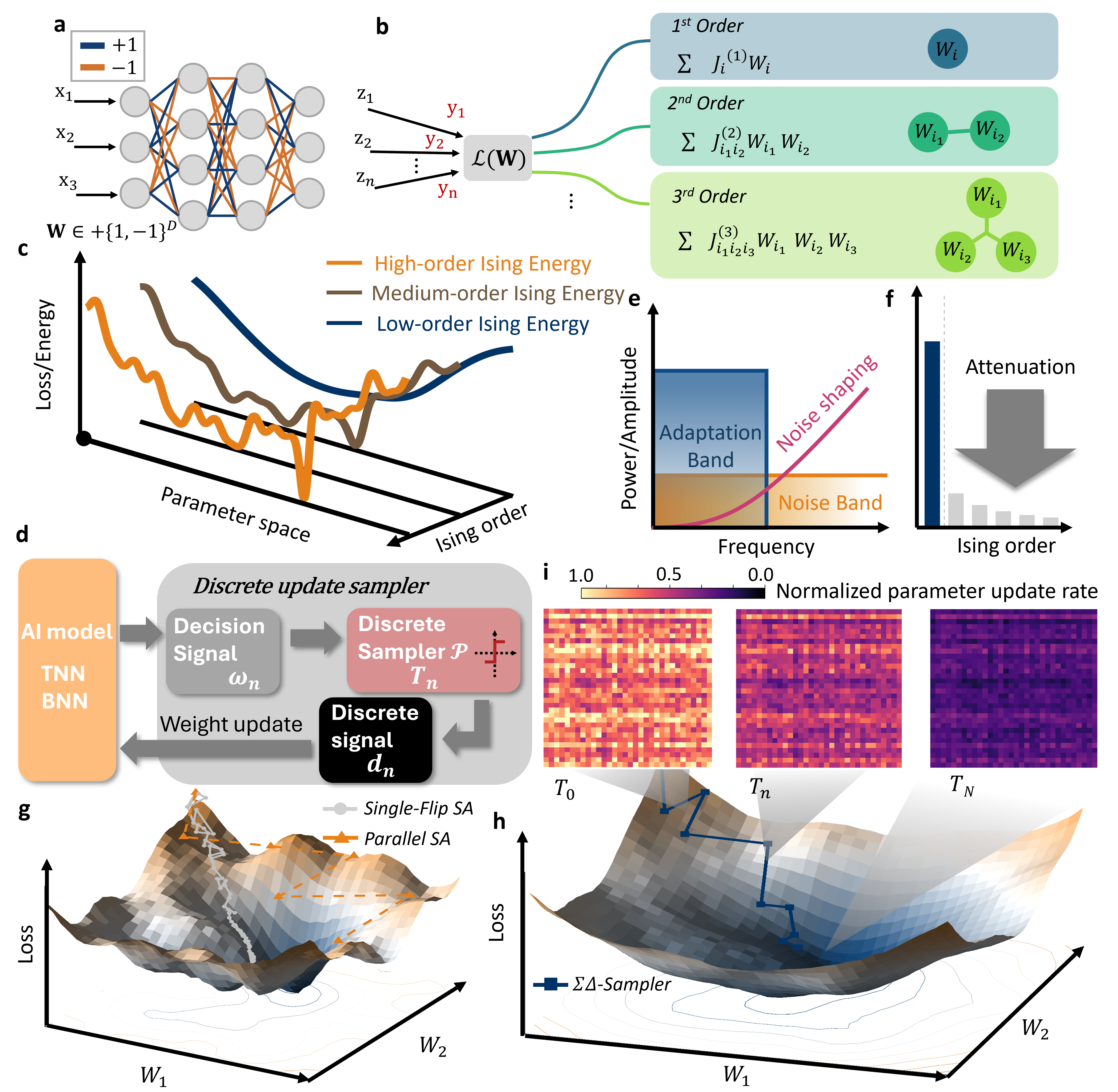}
\caption{\textbf{$\Sigma\Delta$-Sampler for training discrete neural networks.}
(a) A feedforward binary neural network with weights constrained to $\mathbf{W} \in \{-1,+1\}^D$.
(b) Equivalence between discrete parameter neural networks and a hierarchical high-order Ising machine under stationary data distribution $(\mathbf x,\mathbf{y})$.
(c) Conceptual loss/energy profiles for increasing Ising order, where low-order expansions give a smooth landscape and higher-order couplings add frustration, many local minima, and high-frequency components in the gradient space.
(d) A general sampler architecture for training discrete weight neural networks.
(e) The effect of noise-shaping in spectral domain, where the white-noise is pushed out of the low-frequency adaptation band towards high-frequency.
(f) Histogram showing noise-shaping of the Ising decomposition, resulting in the smoother optimization contour that benefits generalization.
(g) Rugged non-convex optimization landscape due to parameter discretizations, where traditional simulated annealing optimizers are slow and do not perform well. 
(h) Optimization trajectory of the $\Sigma\Delta$-Sampler on the smoothened landscape is more stable and computationally scalable.
(i) Parameter update rate maps for a $\Sigma\Delta$-Sampler at different stages of temperature annealing, early($T_0$), mid($T_n$), and late($T_N$), while training a single layer, showing adaptive regularization on the model parameters tracks the smooth optimization contour.
}\label{fig1}
\end{figure}

The Ising model, originally introduced to describe ferromagnetic phase transitions in statistical physics, 
has since become a unifying abstraction for combinatorial optimization~\cite{barahona1982Ising1, Lucas2014Ising2, mohseni2022Ising3}. This has given rise to a rich ecosystem of efficient Ising solvers, including quantum annealers~\cite{Farhi2001QuantumAdiab, Boixo2013EvidenceFQ}, coherent optical Ising machines~\cite{Mwamsojo2023Ocim}, probabilistic-bit~\cite{Singh2024pbit1, Grimaldi2023pbit2} and oscillator-based hardware~\cite{parihar2017vertex, Yin2016EfficientAC}, parallel digital annealing algorithms~\cite{king2022IsingSolver1, maher2024IsingSolver2, cai2020IsingSolver3, isakov2015IsingSolver4}, and neuromorphic architectures~\cite{Chen2025IsingSolver5, Ahsan2026HO}. In its $m$th-order generalization, the Hamiltonian (energy function) of the system takes the form
\begin{equation}
    \mathcal{H}^{(m)}(\mathbf{s}) = -\sum_{c} J^{(m)}_c \prod_{k \in c} s_k,
    \label{eq:ising}
\end{equation}
where $s_k \in \{-1, +1\}$ are binary spins, each clause $c$ is a set of exactly $m$ spin indices, and $J^{(m)}_c$ is the corresponding $m$th-order coupling strength~\cite{Ahsan2026HO}. The appeal of these solvers is that they treat the discreteness of the spin space as a structural feature rather than an obstacle, performing massively parallel stochastic search using only binary states, thermal fluctuations, and local-field arithmetic as computational primitives. This makes these solvers highly efficient on dedicated hardware. To date, Ising solvers have been applied predominantly to combinatorial optimization, where the Hamiltonian is explicitly specified and success is defined as reaching the exact ground state (global minimum) of the energy function. Whether the algorithmic principles underlying these solvers can be leveraged for learning problems remains an open question. In learning,  the goal is not its exact ground state but a solution that minimizes the generalization error. Discrete-weight neural networks are a natural intersection of the two realms. As shown in Fig.~\ref{fig1}a, Binary neural networks (BNNs), which constrain the weights to $\mathbf{W} \in \{-1, +1\}^D$, were originally developed as the extreme point of the model-compression spectrum~\cite{Sze2017CompressionReview, Ding2019Pruning1, Chen2017KD1, Xiao2023PTQ, Jacob2017QuantizationAT}: the special case of 1-bit quantization, applied to the weights and, in fully binarized networks, also to the activations~\cite{Courbariaux2015BinaryConnect, Rastegari2016XNORNet, Liu2023TWN}. Ternary networks extend the weight alphabet to $\{-1, 0, +1\}$, and recent frameworks such as BitNet~\cite{Wang2025Bitnet} enable from-scratch pre-training of large language models (LLMs) with 1-bit (binary) and 1.58-bit (ternary) weights. By replacing multiply--accumulate operations with bitwise XNOR, sign-switching, and popcount instructions, discrete-parameter networks achieve dramatic reductions in both compute and memory, making them compelling candidates for deployment in energy-constrained settings such as embedded sensors, mobile platforms, and edge-intelligence systems~\cite{Chen2026LIM, Li2018EdgeIntell, Abrar2021EdgeIoTReview, sunny2021BNNaCC1, andri2016BNNaCC2, conti2018BNNaCC3, vo2022BNNaCC4}.

The structural correspondence between Ising models and discrete-weight networks is illustrated in Fig.~\ref{fig1}b. Consider a network of arbitrary architecture with network output $\mathbf{z}$, whose inputs $\mathbf{x}$ and targets $\mathbf{y}$ are drawn from a stationary data distribution. The training loss $\mathcal{L}$ is then a function of the discrete weights $\mathbf{W}$ alone, that is, a real-valued function on the finite set $\{-1,+1\}^D$ of $2^D$ weight configurations. For every subset $S \subseteq \{1, \dots, D\}$ of weight indices, define the parity monomial $\prod_{i \in S} W_i$. These $2^D$ monomials form an orthonormal basis (the Walsh--Fourier basis) for the space of real-valued functions on $\{-1,+1\}^D$, so the loss admits a unique and exact decomposition
\begin{equation}
    \mathcal{L}(\mathbf{W}) \;=\; \sum_{m=0}^{D}\ \sum_{|S|=m} J^{(m)}_S \prod_{i\in S} W_i,
    \label{eq:bnn-ising}
\end{equation}
where $J^{(m)}_S$ is the $m$th-order coupling among the $|S| = m$ weights in $S$, and the $m=0$ term is a constant offset. The couplings are determined implicitly by the loss function, the network architecture, and the data distribution, but the decomposition itself is architecture-agnostic, and its derivation is given in Methods Section~\ref{sec_methods_walsh}. Identifying the discrete weights $\mathbf{W}$ with the spins $\mathbf{s}$, and absorbing the sign convention of Eq.~\eqref{eq:ising} into the couplings, the loss becomes a hierarchical sum of Ising Hamiltonians of every order, $\mathcal{L}(\mathbf{W}) = \sum_{m=0}^{D} \mathcal{H}^{(m)}(\mathbf{W})$. The same construction extends to ternary weights. Discrete-weight networks are therefore precisely the model class on which an Ising solver can operate, with one caveat that in practice, the loss is evaluated on mini-batches, so the effective couplings are stochastic estimates that fluctuate from one iteration to the next and are only quasi-stationary. One approach for training the network is using post-training quantization~\cite{Xiao2023PTQ,Shen2021PTQ1}, in which $\mathcal{L}$ is optimized in the continuous domain and the final floating-point solution is projected onto the discrete-space. However, this procedure has been shown to be sub-optimal compared with quantization-aware training in the extreme-low-bit domain~\cite{Jacob2017QuantizationAT,Krishnamoorthi2018QAT2}. Moreover, our focus is on methods that enable training and fine-tuning directly in the discrete domain on edge devices. 

Irrespective of the choice of the Ising solver or BNN training algorithm, any optimizer that iteratively minimizes $\mathcal{L}(\mathbf{W})$ over a discrete set $\Omega$ (e.g., $\{-1,+1\}^D$ or $\{-1,0,+1\}^D$) can be written as
\begin{equation}
    \mathbf{W}^n \;=\; \mathbf{W}^{n-1} + \mathbf{d}^{n}, \qquad \mathbf{W}^n, \mathbf{W}^{n-1} \in \Omega,
    \label{eq:basic-update}
\end{equation}
where $n$ is the iteration index and the update vector $\mathbf{d}^{n}$ is constrained such that $\mathbf{W}^n$ remains in $\Omega$. Note that for a deterministic flip of the $\text{i}^{th}$ spin or weight ($W_i^n \rightarrow -W_i^{n-1}$), the $\text{i}^{th}$ element of the update vector $d_i^{n} = -2W_i^{n-1}$. 
The update rule in Eq.~\eqref{eq:basic-update} can be viewed as drawing $\mathbf{d}^n$ at each iteration from an underlying distribution as shown in Fig.~\ref{fig1}d

\begin{equation}
    \mathbf{d}^n \sim \mathcal{P} \!\left(\mathbf{\boldsymbol\omega}^{n},\, \mathbf{W}^{n-1},\, T^n\right)  \label{eq:sampling}
\end{equation}


Here $\mathcal{P}$ denotes a general sampling process parameterized by a decision signal $\mathbf{\omega}^{n}$, the current weights $\mathbf{W}^{n-1}$, and a temperature $T^n$. In simulated annealing (SA), each weight update (spin flip) is accepted according to a Boltzmann criterion~\cite{Kirkpatrick1983}, with the decision signal $\mathbf{\omega}^n=\Delta \mathcal{H}^n$ denoting the change of the Ising Hamiltonian as a result of the spin state configuration change, while the temperature is lowered according to an annealing schedule. Both ingredients map naturally onto dedicated annealing hardware~\cite{Chen2025IsingSolver5, Ahsan2026HO} and are closely related to quantum annealing~\cite{Boixo2013EvidenceFQ}. The temperature controls the degree of randomness in the sampling process, with $T^n = 0$ corresponding to deterministic sampling. The straight-through estimator (STE) corresponds to this zero-temperature limit, and it constitutes the dominant quantization-aware training paradigm~\cite{Bengio2013STE, Neftci2019SurrogateGL, Courbariaux2015BinaryConnect, Hubara2016BNN}. STE optimizer samples the discrete update decision by maintaining a full-precision latent copy $\hat{\mathbf{W}}$ of every discrete weight as the decision signal $\mathbf{\omega}^n$, re-quantizes $\mathbf{W} = \mathrm{sign}(\hat{\mathbf{W}})$ for the forward pass, and applies the surrogate gradient $\mathbf{g}^n$ in updating $\hat{\mathbf{W}}$ through conventional gradient-based optimizers with the predominant one being AdamW~\cite{kingma2017AdamOpt, zhu2024scalable}. Numerous refinements to threshold functions, clipping ranges, and binarization schedules have been proposed~\cite{Liu2020ReActNet, Martnez2020TrainingBN, Qin2019ForwardAB, Liu2018BiRealNet, Liu2023TWN}.

However, to generalize well, any weight update rule in Eq.~\eqref{eq:basic-update} must overcome the challenges revealed by the multi-order decomposition of the loss function, as shown in Fig.~\ref{fig1}c. The low-order terms form a smooth envelope with broad basins and well-defined descent directions, because they are dominated by a small number of strong, slowly varying couplings. The high-order terms, in contrast, superimpose fine-grained frustration onto that envelope where exponentially many weak clauses impose competing constraints that cannot all be satisfied simultaneously, producing a proliferation of local minima~\cite{Derrida1981LocalMinHOIsing, Stariolo2020HOIsingTrap}. Consequently, along the temporal trajectory of any iterative optimizer (such as SA), the contribution of the high-order terms to the update signal fluctuates rapidly (high-frequency), whereas the contribution of the low-order envelope varies slowly (low-frequency). Higher-order frustration is a principal, but not the only, source of these fluctuations. Weight quantization introduces its own error, mini-batch sampling perturbs the couplings from one step to the next, and the stochastic transitions of the sampler add further randomness. Although these sources differ in origin, they share a spectral signature. None of these signals are correlated with the slowly varying descent direction, and each, to a first approximation, exhibits a white-noise (or blue-noise) spectrum across the iteration band. The gradient-descent component, by contrast, arises from the low-order couplings, which change slowly as the optimizer traverses the loss surface and therefore occupy the low-frequency portion of the weight-update trajectory. The STE optimizer omitted the challenges imposed by parameter discretization by tracking the complete continuous counterparts through its decision signal $\boldsymbol\omega^n$ regardless of the spectral properties of the signal. On the other hand, because signal and noise are separated in frequency rather than in amplitude, the natural remedy is to reshape the noise spectrum rather than to suppress the noise outright. 

In this work, we propose a new mechanism for generating the updates $\mathbf{d}^n$, driven by the error between the cumulative gradient and the cumulative update, which we call $\Sigma\Delta$ sampling and can be expressed as:
\begin{equation}
    \mathbf{d}^n  \sim \mathcal{P} \!\left(\left|\sum_{k=1}^{n}\mathbf{g}^k - \sum_{k=1}^{n-1}\mathbf{d}^k \right|,\, \mathbf{W}^{n},\, T^n\right). \quad\text{($\Sigma\Delta$ Sampling)}
    \label{eq:basic-SD}
\end{equation}
Because the sampler is driven by the accumulated error, the low-pass-filtered gradient trajectory and the low-pass-filtered discrete-update trajectory are forced to remain close to each other. Equivalently, $\Sigma\Delta$-Sampler pushes the white-noise energy introduced by quantization, Ising frustration, and mini-batching out of the low-frequency adaptation band and toward the high-frequency, a mechanism known as noise-shaped learning~\cite{Chakrabartty2013NoiseShaped}. Therefore, $\Sigma\Delta$-Sampler lowers the noise floor within the frequency band occupied by the gradient-descent signal as shown in Fig.~\ref{fig1}e. Since the fluctuating component originates partly from the high-order couplings, the sampler exposes the low-order structure at a higher signal-to-noise ratio demonstrated in Fig.~\ref{fig1}f, and the optimizer descends along the smooth low-order envelope rather than tracking the frustration layered on top of it. Note that noise-shaping principles also form the core of high-resolution $\Sigma\Delta$ data converters~\cite{Candy1991SD}, where quantization noise is moved out of the signal band of interest. The $\Sigma\Delta$-Sampler described in Eq.~\eqref{eq:basic-SD} can be implemented iteratively as
\begin{align}
    \mathbf{\boldsymbol\omega}^n &= \mathbf{\boldsymbol\omega}^{n-1} + \mathbf{g}^n - \boldsymbol{\beta}^n \odot \mathbf{d}^{n-1}, \quad\text{(Integrator)} \label{eq:intro_sd_int}\\
    \mathbf{d}^n & \sim \mathcal{P}\!\left(\mathbf{\boldsymbol\omega}^n,\, T^n\right), \quad\text{(Sampling)} \label{eq:intro_sd_quant}
\end{align}
where $\odot$ denotes a Hadamard product, $\mathbf{\omega}^n$ is the integrator state/$\Sigma\Delta$ decision signal, $\boldsymbol{\beta}^n$ is a dynamic feedback-gain, and $\mathcal{P}(\,\cdot\,, T^{n})$ is a sampling function detailed in Methods Section~\ref{sec_methods_sigma_delta}. As in other annealing approaches, the temperature $T^{n}$ controls the degree of randomness in the sampling process. The first-order recursion in Eq.~\eqref{eq:intro_sd_int} and~\eqref{eq:intro_sd_quant} ensures that $\mathbf{\omega}^n$ remains bounded, hence ensuring the stability of the sampling process. The integrator state $\mathbf{\omega}^n$ varies smoothly during training and preserves the low-frequency content of the loss landscape that encodes the descent direction. Whereas the per-batch gradient $\mathbf{g}^n$ is dominated by high-frequency fluctuations arising from the many local minima of the discrete landscape.

In Fig.~\ref{fig1}g, different optimization dynamics are illustrated conceptually using a
highly non-convex discrete landscape in ~\cite{QIN2020BNNsurvey, Baldassi2020BNNrugged1, zhou2022COPBNN}, where we have showed the speculative trajectories of \emph{Single-flip SA} and \emph{Parallel SA} optimizers in comparison to the dynamics of $\Sigma\Delta$-Sampler in Fig.~\ref{fig1}h. \emph{Single-flip SA} proposes one spin flip per iteration; it provably converges to the ground state under a sufficiently slow cooling schedule, but it explores the configuration space sequentially~\cite{Kirkpatrick1983, Hajek1988OptimalCooling, geman1984stochastic}. \emph{Parallel SA} proposes flips for all spins simultaneously; it scales naturally to high-dimensional networks, but cannot provide convergence guarantee as its single-flip variant. Single-flip SA (gray trajectory) is thermally stable but tends to become trapped among the many local minima of the rugged landscape, and its computational cost becomes intractable at the scale of modern deep networks. Unconstrained parallel SA (orange trajectory) makes long-range moves, but because simultaneous flips are decided from noisy local fields that ignore their mutual interactions, the joint energy change of a multi-spin flip is uncontrolled, causing the optimizer to overshoot and diverge. Fig.~\ref{fig1}h shows how the noise-shaping transforms this rugged surface into a smoothed effective landscape on which the $\Sigma\Delta$-Sampler descends stably and long-range parallel transitions remain energetically well behaved. $\Sigma\Delta$-Sampler thus inherits the stability of single-flip SA without its per-coordinate, per-iteration cost, and the scalability of parallel SA without its divergence on rugged landscapes. In other words, a single mechanism closes both the \emph{dynamics gap} (stability versus scalability) and the \emph{objective gap} (ground state versus generalization). In the Results section, we show that $\Sigma\Delta$-Sampler converges substantially faster than standard SA approaches and some benchmark STE approaches. Because $\mathbf{\omega}^n$ acts as a per-parameter signal-to-noise readout of the smoothed landscape, the weight-flip activity (changes in weights) are concentrated where the landscape is steep and decays as coordinates settle into a basin. This produces the spatially and temporally structured update-rate maps of Fig.~\ref{fig1}i at early ($T^0$), intermediate ($T^n$), and late ($T^N$) training stages. 
The per-parameter update budget therefore self-regulates without hand-tuned per-parameter schedules, yielding an emergent, landscape-aware regularization that is analogous to an adaptive learning rate in discrete parameter space.

A further benefit of performing the entire optimization in the discrete domain is best seen in contrast with STE. STE is practical precisely because conventional optimizers acting on the full-precision latent weights operate on the comparatively benign continuous loss landscape of deep networks. 
The $\Sigma\Delta$-Sampler, in contrast, modulates the discrete states directly. Neither the latent weights nor the first- and second-moment optimizer states attached to them~\cite{kingma2017AdamOpt} are needed, so the training memory footprint tends towards the inference memory footprint. There have been other attempts to remove the latent weights from the training of binary networks, for example by replacing them with a moving average of the gradient~\cite{Helwegen2019LatentDoNotExist} which suggests that a full-precision latent copy is not necessary for good performance. The combination of a self-regulating update budget and a near-inference-scale training footprint is important for edge and embedded deployment, where models must keep adapting to new data (non-stationary operating conditions) under tight memory and energy budgets~\cite{Hayes2020REMIND, Hayes2022OnlineCL, pellegrini2020latentreplayrealtimecontinual, lin2024ondevicetraining256kbmemory, Cai2020TinyTL, Lin2020MCUNet}. Note that for edge and embedded AI applications, carrying a full-precision training stack defeats the very purpose of quantizing the parameters. In the Results section, we first validate the framework on a planted QUBO benchmark whose exact minimizer is known by construction. We then apply the framework to standard vision benchmarks (CIFAR-10 and ImageNet-1k), class-incremental continual learning, and language-model pre-training at the scale of billions of parameters. Across these settings, $\Sigma\Delta$-Sampler attains convergence dynamics close to those of state-of-the-art full-precision baselines while operating strictly in the discrete parameter space throughout, with a substantially reduced training memory footprint that makes on-device continual adaptation practical.

\section{Results}\label{sec2}


Our empirical investigation progresses from controlled synthetic settings with known ground truth to full-scale neural network benchmarks. First, we conduct a planted Quadratic Unconstrained Binary Optimization (QUBO) experiment to isolate the dynamics of the $\Sigma\Delta$ noise-shaping loop under exact binary minimizers (Section~\ref{sec_res_qubo}). Next, we evaluate the $\Sigma\Delta$-Sampler on CIFAR-10 image classification (Section~\ref{sec_res_benchmarks}), benchmarking discrete-space training against STE baselines in both standalone and continual learning regimes (Section~\ref{sec_res_continual_adaptation}). Finally, we examine the hardware energy and memory implications of our discrete-throughout-training design on larger datasets (ImageNet-1K) and large language models (LLMs) (Section~\ref{sec_res_scaling} and ~\ref{sec_res_hw}).

\begin{figure}[h!]
\centering
\includegraphics[width=0.99\textwidth]{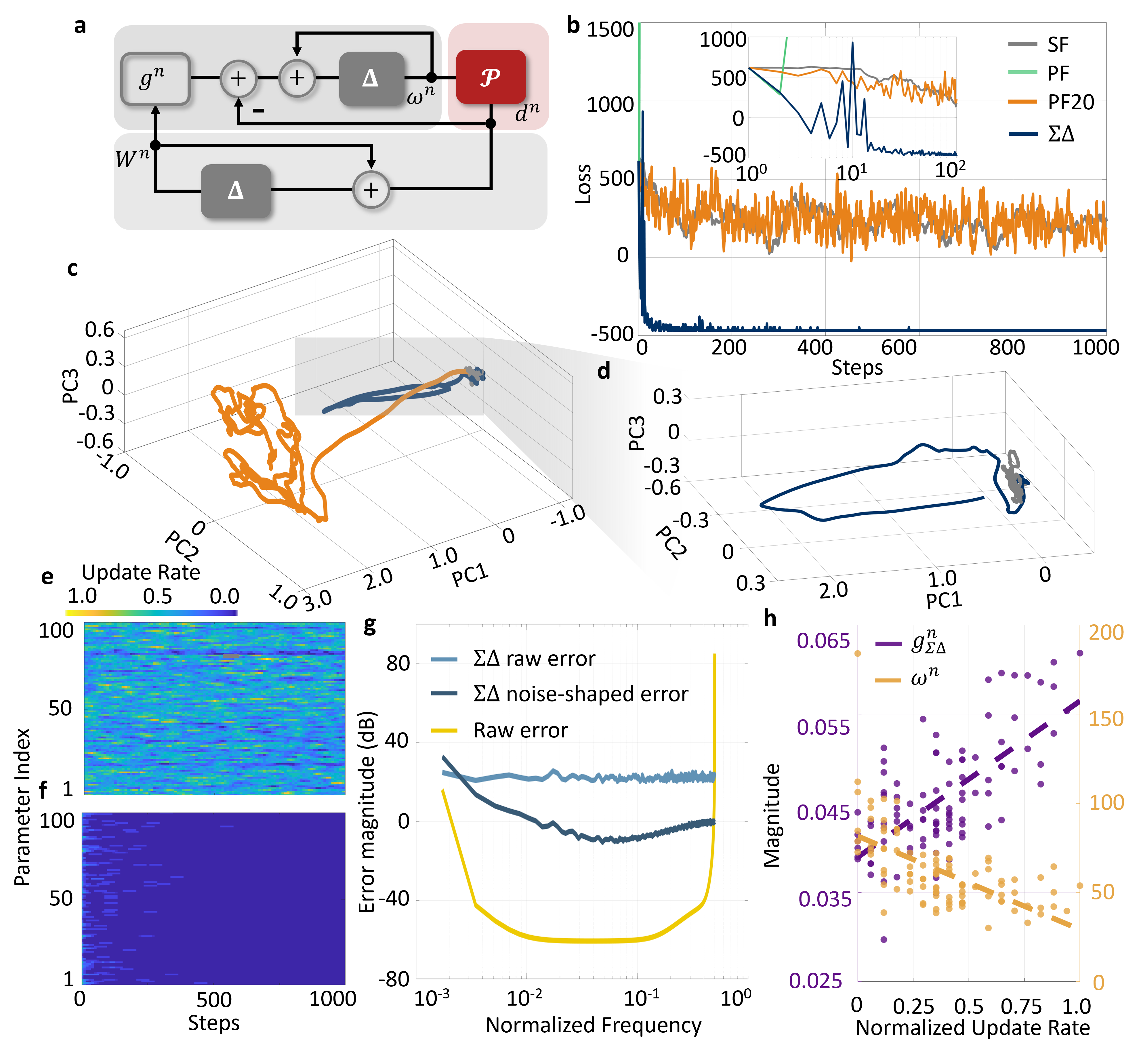}
\caption{\textbf{Spectral and dynamic analysis of $\Sigma\Delta$-Sampler on a planted Quadratic Unconstrained Binary Optimization (QUBO) problem.} {(a)} Block diagram of the $\Sigma\Delta$-Sampler update rule. {(b)} Optimization plot showing convergence towards the planted global optimum $\mathbf{W}^*$ over $1{,}000$ steps for four different optimizers: \textbf{SF} (single-flip SA), \textbf{PF} and \textbf{PF-20} (parallel-flip SA at $100\%$ and $20\%$ flip rates respectively), and \textbf{$\Sigma\Delta$}-Sampler. {(c)} Principal component visualization of the optimization trajectories, with {(d)} showing the convergence on a shorter time-scale. {(e), (f)} Temporally windowed normalized parameter update rate of \textbf{PF} SA and $\Sigma\Delta$-Sampler showing smoothing of the parameter update trajectories due to noise shaping. (g) Power spectra of the quantization errors (difference in the decision signal vs. update signals) in different optimizers, showing the effect of noise-shaping. (h) Linear extrapolation of average $|g^n|$ and $|\omega^n|$ against overall parameter update rate during the optimization process.}\label{fig2} 
\end{figure}

\subsection{Validation of $\Sigma\Delta$-Sampler on small-scale QUBO}\label{sec_res_qubo}

To isolate the mechanisms of $\Sigma\Delta$ noise shaping in a controlled setting, four algorithms were evaluated on a planted Quadratic Unconstrained Binary Optimization (QUBO) instance whose ground state is known by construction. We sample a uniformly random binary upper-triangular coupling matrix $\mathbf{Q} \in \{-1,+1\}^{D \times D}$ and shift its diagonal by $\lambda_{\max}\mathbf{I}$, where $\lambda_{\max}$ is chosen strictly larger than the spectral radius of $\mathbf{Q}$ to ensure positive definiteness. The QUBO loss is given by
\begin{equation}
    \mathcal{L}(\mathbf{W}) \;=\; \tfrac{1}{2}\,\mathbf{W}^\intercal\,\mathbf{Q}\,\mathbf{W} - \mathbf{b}^{\intercal}\mathbf{W},
    \label{eq_planted_qubo}
\end{equation}
where the bias $\mathbf{b}=\mathbf{Q}\mathbf{W}^*$ is set from a uniformly sampled planted binary optimum $\mathbf{W}^*\in \{-1,+1\}^D$. By construction $\mathbf{W}^*$ is the globally optimal discrete minimizer of $\mathcal{L}$ on the binary hypercube. The experiments shown in Fig.~\ref{fig2} use $D=100$. Under this setup, we compare three simulated annealing (SA) configurations with $\Sigma\Delta$-Sampler, all sharing the same hyperbolic temperature schedule $T^n = T_0 / (T_0 + n/\tau)$:
\begin{itemize}
    \item \textbf{SF} (Single-Flip SA): One coordinate $i$ is selected uniformly at random per iteration and flipped with probability $\exp(-|g^n_i| / T^n)$.
    \item \textbf{PF} (Parallel-Flip SA): All coordinates evaluate flips synchronously at each iteration, with each per-coordinate decision driven by its raw per-iteration gradient.
    \item \textbf{PF-20} (Partial Parallel-Flip SA): Same as PF, but with the per-iteration flip fraction restricted to 20\%, throttling PF's aggressiveness while retaining its parallelism.
    \item \textbf{$\Sigma\Delta$} ($\Sigma\Delta$-Sampler): Parallel decisions driven by the $\Sigma\Delta$-Sampler (Figure~\ref{fig2}a). The error between the discrete update signal $\mathbf{d}^n$ and the raw gradient $\mathbf{g}^n$ is accumulated in the integrator $\boldsymbol{\omega}^n$, which is then fed into the sampler $\mathcal{P}$ to produce $\mathbf{d}^n$.
\end{itemize}

Figure~\ref{fig2}b shows that \textbf{SF} decreases the loss with persistent oscillation but never reaches the global minimum $\mathbf{W}^*$ within the recorded time window, whereas \textbf{PF} diverges immediately without recovering, and \textbf{PF-20} suppresses the most violent oscillations of \textbf{PF} without restoring convergence. The geometric signature of these failure modes is captured in the joint Principal Component Analysis (PCA) projection of the discrete-update streams (Figure~\ref{fig2}c; construction detailed in Methods Section~\ref{sec_methods_pca}). \textbf{PF} is omitted from the projection because its trajectory dominates the principal components without yielding informative insights about the other variants. \textbf{PF-20} sweeps a broad volume of the parameter space through large, incoherent excursions that double back on themselves, while \textbf{SF} (magnified in the inset of Figure~\ref{fig2}d) remains confined to a microscopic volume. The root cause of both failures is the absence of temporal memory in the per-iteration gradient, which is dominated by the local structure of the non-convex landscape, leading to significant oscillation.

In contrast, the $\Sigma\Delta$-Sampler recovers $\mathbf{W}^*$ exactly in Figure~\ref{fig2}b with the inset highlights an initial \emph{high-flip-rate} exploration phase followed by rapid convergence into the basin within $\sim 100$ steps. In the PCA projection in Figure~\ref{fig2}c,d, the $\Sigma\Delta$-Sampler traces a broad trajectory through the joint PC space before converging onto the planted solution, effectively converting uncoordinated and parallel random walk into coherent macro-steps along the noise-shaped loss-contour. The per-parameter origin of this behavior is demonstrated in Figure~\ref{fig2}e,f, which displays time-binned, normalized flip activity per parameter across optimization for \textbf{PF-20} and $\Sigma\Delta$, respectively. \textbf{PF-20} produces broadband, undifferentiated activity across all coordinates and iterations, oscillating against per-iteration noise regardless of the location on the loss surface. Under the same temperature schedule, $\Sigma\Delta$-Sampler produces a temporally regularized update map consistent with Figure~\ref{fig1}i, where per-coordinate activity is modulated proportionally to the local gradient and then decays smoothly toward zero near the basin.

The spectral signature in Figure~\ref{fig2}g identifies the mechanism underlying these different temporal trajectories. Each curve depicts the power spectral density (PSD) of the residual between a loop's decision signal ($\mathbf{g}^n$ or $\boldsymbol{\omega}^n$) and its realized flip output $\mathbf{d}^n$ (representing the per-step quantization error). For the open-loop raw error (yellow, $\mathbf{g}^n_{PF}-\mathbf{d}^n_{PF}$) extracted from the \textbf{PF} optimizer, the error spectrum is dominated by a peak at the Nyquist frequency which is a hallmark of a period-2 limit cycle in which each step's flips are immediately reversed on the subsequent step, expending the iterations on oscillations within the loss-lanscape rather than descent. The rise in energy-content at low frequencies represents the smooth component of the gradient embedded within the error signal, indicating that the landscape's descent contour is discarded by the flips. When the useful signal is buried inside a Nyquist-dominated error, the loop fails to converge. 

Conversely, the same residual for the $\Sigma\Delta$-Sampler, $\mathbf{g}_{\Sigma\Delta}^n - \mathbf{d}_{\Sigma\Delta}^n$ (light blue), exhibits no low-frequency structure and the systematic descent content of the gradient is fully absorbed within the parameter trajectory, leaving uncorrelated per-step error for the accumulator $\boldsymbol{\omega}^n$ to integrate. As a result, the persistent descent component accumulates coherently while transient noise averages out. The accumulator residual (dark blue, $\boldsymbol{\omega}^n-\mathbf{d}^n_{\Sigma\Delta}$) then exhibits the characteristic noise-shaping signature which is a V-shaped spectrum whose low-frequency branch represents the preserved learning signal and whose high-frequency branch rising toward Nyquist reflects the first-order-shaped quantization noise of the flip output, consistent with the noise transfer function (Methods Section~\ref{sec_methods_sigma_delta}).

Figure~\ref{fig2}h characterizes this population-level mechanism by plotting the time-averaged gradient magnitude $|\mathbf{g}_{{\Sigma\Delta}}|$ and accumulator magnitude $|\boldsymbol\omega|$ against the normalized per-parameter update rate across all $D=100$ parameters. The two quantities exhibit opposing linear trends. Coordinates with large $|\mathbf{g}_{{\Sigma\Delta}}|$ flip frequently and maintain small $|\boldsymbol\omega|$ which signifies a low signal-to-noise ratio (SNR) regime where the integrator discharges rapidly after each flip because of noise embedded in gradient signal with large amplitude. Conversely, coordinates with small $|\mathbf{g}_{{\Sigma\Delta}}|$ flip rarely and accumulate larger $|\boldsymbol\omega|$ which signifies a high-SNR regime where a clear decent direction builds up in the decision signal. While the noise-shaping spectrum of Figure~\ref{fig2}g applies identically to every parameter, the local SNR set by $|\mathbf{g}_{{\Sigma\Delta}}|$ dictates the individual update frequency, causing the update budget to self-regulate dynamically. This confirms at the population level the per-parameter SNR mechanism proposed in the introduction and illustrated in Figure~\ref{fig1}i.

\subsection{$\Sigma\Delta$-Sampler results on standard vision benchmarks}\label{sec_res_benchmarks}
\begin{figure}[h!]
\centering
\includegraphics[width=0.95\textwidth]{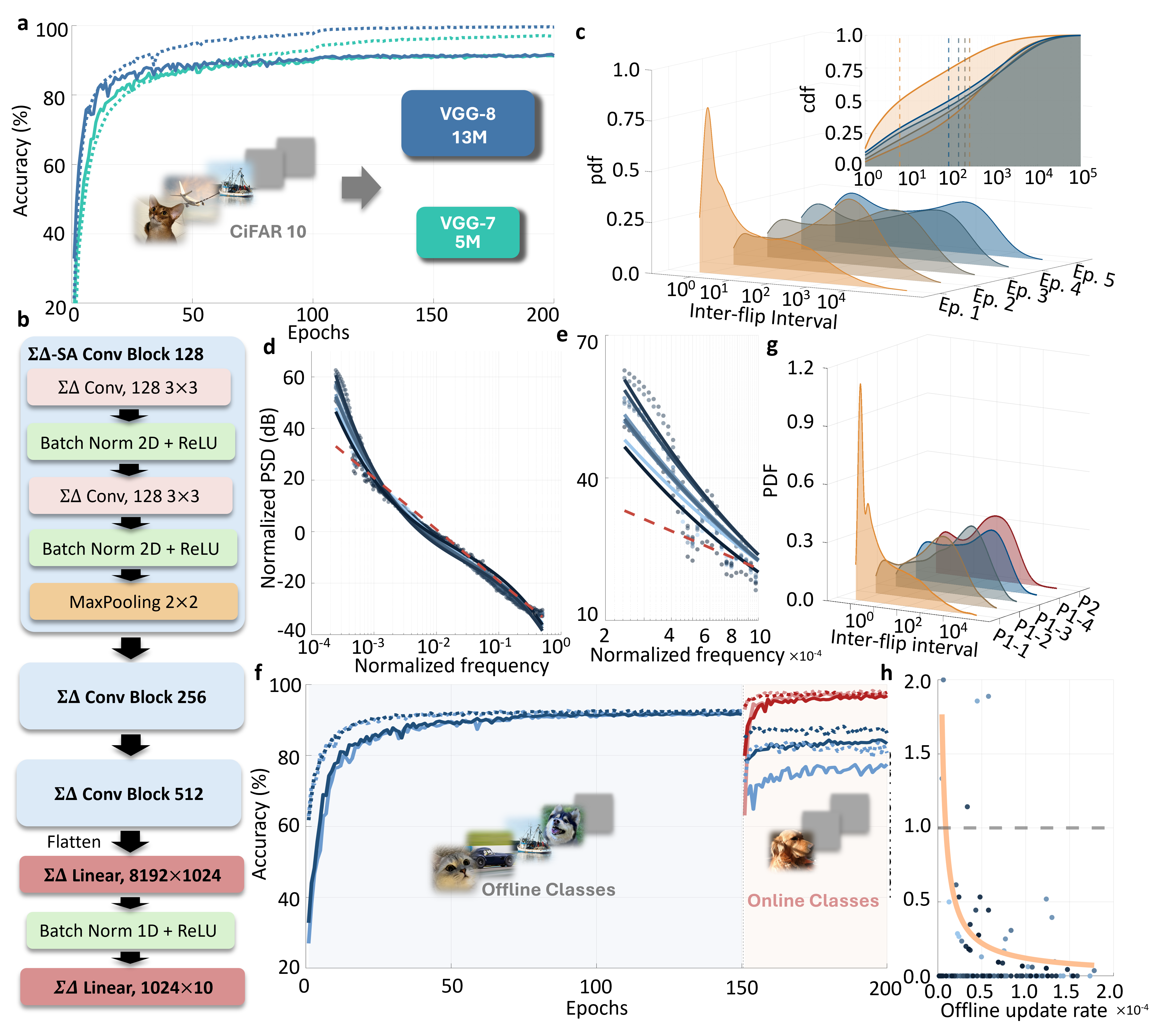}
\caption{\textbf{$\Sigma\Delta$-Sampler training dynamics on CIFAR-10 and the spectral signature of $\Sigma\Delta$ noise-shaping.} (a) Train (dashed) and test (solid) accuracy of two ternary networks trained end-to-end with the $\Sigma\Delta$-Sampler pipeline on CIFAR-10: VGG-7 and VGG-8. (b) Block diagram of the used convolution model. (c) Per-epoch probability density function (pdf) of per-parameter inter-flip intervals where color denotes training stage from early (orange) to late (blue) epochs; inset shows the corresponding cumulative distribution function (cdf), where the dashed line marks the median. (d) Normalized empirical power spectral density of the per-parameter $\Sigma\Delta$ integrator state $\omega^n_j$ estimated over a randomly sampled cohort of weights, compared against a $1/f^2$ reference spectrum (marked in red dashed line) overlaid, with (e) showing the low-frequency zoom of the same PSD. (f) Validation accuracy for $\Sigma\Delta$-Sampler (solid trajectory) and full-precision counterparts (dashed trajectory) in continual-learning configuration, where the different color codes represent different offline-class replay fractions. (g) Per-phase probability density of per-parameter inter-flip intervals across the offline phase (orange to blue) and the online phase (red). (h) Per-parameter reactivation ratio plotted against the phase-1 update rate.}\label{fig3}
\end{figure}

Having validated the behavior of $\Sigma\Delta$ noise-shaping under the controlled QUBO setting (Section~\ref{sec_res_qubo}), we evaluate the $\Sigma\Delta$-Sampler as a practical training algorithm for deep quantized networks on standard image classification benchmarks, and to characterize it dynamics in neural network training. Figure~\ref{fig3}a presents the training (dashed) and validation (solid) accuracy trajectories of VGG-7 and VGG-8, both trained end-to-end with the $\Sigma\Delta$-Sampler pipeline on CIFAR-10. Figure~\ref{fig3}b illustrates the modified VGG-7 convolutional architecture adapted from~\cite{Liu2023TWN} while VGG-7 differs by omitting the $8192\times1024$ linear layer. The two models achieve similar final test accuracies, with the larger model outperforming the smaller variant by less than $1\%$ at convergence. As reported in Table~\ref{tab1}, both models retain $>99\%$ of the accuracy of the full-precision baseline on CIFAR-10, despite the parameter reduction in the 5M-parameter VGG-7 variant. Furthermore, our approach achieves solution quality comparable to existing discrete neural network baselines. Ternary Weight Networks (TWN) reported a top test accuracy of $92.56\%$ in their original work~\cite{Liu2023TWN}, and we obtained a top accuracy of $91.56\%$ by executing the training script from their original codebase following the configuration proposed in the literature. However, the default TWN model configuration as proposed in the original literature includes a full-precision linear classifier head, whereas our model is fully discrete. For fairness of comparison, we performed additional training runs with fully discrete TWN model and last full-precision linear classifier attached to $\Sigma\Delta$-Sampler model. As shown in Table~\ref{tab1}, the full-precision layer contributed to ~$0.1\%$ final accuracy, where TWN performs better with FP classifier while $\Sigma\Delta$ sampler performs worse. 


Table~\ref{tab1} also details the per-epoch training time for all runs executed on identical hardware (NVIDIA H100 80GB). On average, the $\Sigma\Delta$-Sampler requires $13.4$\,s/epoch and $11.9$\,s/epoch in ternary mode across the two model variants, and $9.0$\,s/epoch in binary mode, compared to $9.2$\,s/epoch for full-precision training. On a per-epoch basis, ternary annealing introduces a $1.45\times$ overhead, whereas binary annealing is essentially compute-neutral ($0.97\times$). However, because the $\Sigma\Delta$-Sampler incorporates initial warmup epochs, total training runtime is higher due to the extended schedule ($200$ epochs vs. $150$ epochs). In contrast, the TWN pipeline matches the total epoch count ($150$) of its full-precision counterpart as reported in~\cite{Liu2023TWN}, but requires $247.8$\,s/epoch---an $18.5\times$ slowdown relative to our ternary runtime. Finally, for completeness, Supplementary Information Section~\ref{sec_supp_gpupower} reports measured GPU power draw across optimizers under identical training setups and we attribute the observed energy variations primarily to algorithmic implementation differences.

\begin{table}[h]
\centering
\caption{Final test accuracy and time to presented solution on CIFAR-10 for full-precision baseline model, Ternary Weight Networks (TWN) model~\cite{Liu2023TWN} and the proposed approach $\Sigma\Delta$-Sampler on the VGG CNN architecture (13M). *The accuracy reported in the original TWN work is $92.56\%$}\label{tab1}
\footnotesize
\begin{tabular*}{\textwidth}{@{\extracolsep{\fill}}lllr}
\toprule
Method & Weight precision & Accuracy (\%) & Time to solution (h)\\
\midrule
FPWN (fp baseline) & fp32 & 92.88 & 0.38\\
TWN FP classifier & ternary $\{-1,0,+1\}$ & 91.56* &10.32\\
TWN discrete& ternary $\{-1,0,+1\}$ & 91.47 &10.85\\
$\Sigma\Delta$-Sampler FP classifier & ternary $\{-1,0,+1\}$ & 92.16 &0.69\\
$\Sigma\Delta$-Sampler discrete& ternary $\{-1,0,+1\}$ & 92.20 &0.73\\
$\Sigma\Delta$-Sampler, $\sim$5M parameters & ternary $\{-1,0,+1\}$ & 91.88 &{0.67}\\
$\Sigma\Delta$-Sampler & binary $\{-1,+1\}$ & 91.89 &0.50\\
\bottomrule
\end{tabular*}
\end{table}

For the remainder of this section, we analyze the optimization dynamics of the $\Sigma\Delta$-Sampler when training the VGG-8 model. The temporal structure of the per-parameter update stream is quantified at the population level in Figure~\ref{fig3}c, which illustrates the distribution of per-parameter inter-flip intervals across training stages. The shift in distribution shape over time reflects a uniform change in flip activity across all parameters. Figure~\ref{fig3}c plots the probability density function (pdf) along a logarithmic interval axis. In the earliest stage, the distribution is heavily concentrated at short intervals, indicating that most parameters flip at every iteration. Later stages broaden into a roughly uniform distribution with no single dominant peak. Because each flip corresponds to a write operation on the persistent weight tensor, the systematic lengthening of inter-flip intervals throughout training directly reduces off-chip memory traffic, of which the implication will be discussed further in Section~\ref{sec_res_hw}. The inset shows the corresponding cumulative distribution function (cdf), with dashed vertical lines marking the median interval for each stage. Notably, the median exhibits non-monotonic behavior over training time: the second stage exhibits the highest median, while subsequent stages drift back toward shorter intervals without a monotonic trend.

Figures~\ref{fig3}d and~\ref{fig3}e provide spectral evidence for the noise-shaping effect within the $\Sigma\Delta$ accumulator signal. For a randomly sampled cohort of weights across all layers, we record the per-iteration accumulator sequence $\boldsymbol\omega^n$ throughout training and compute its empirical power spectral density (PSD), normalized such that the integrated log-spectral energy is matched across layers and against the open-loop baseline. Figure~\ref{fig3}d displays the resulting PSD over the full frequency spectrum, while Figure~\ref{fig3}e zooms into the low-frequency band. Here, the color gradient denotes layer depth, ranging from light (shallow) to dark (deep). The red dashed line represents a $1/f^2$ reference spectrum which represents the asymptotic spectrum of a pure integrator driven by white noise, corresponding to the open-loop limit of the $\Sigma\Delta$ recursion (Eq.~\eqref{eq:intro_sd_int}) with zero feedback gain. This reference models the spectral profile that would arise if the integrator merely accumulated unshaped noise. In contrast, the measured closed-loop PSD is everywhere steeper than the $1/f^2$ baseline. In the low-frequency view as in Figure~\ref{fig3}e, the closed-loop curve bends away from $1/f^2$ toward an even steeper roll-off. Under log-spectral energy normalization, this steeper slope confirms a structural reallocation of spectral mass from the high-frequency regime into the low-frequency band where the smooth descent signal resides. Parameter update decisions are directly driven by the $\Sigma\Delta$-shaped accumulator $\boldsymbol\omega^n$, wherein closed-loop feedback suppresses high-frequency quantization noise and concentrates energy in the low-frequency band to support stable gradient descent.

\subsection{$\Sigma\Delta$-Sampler for continual adaptation}\label{sec_res_continual_adaptation}
\begin{table}[h]
\centering
\caption{CIFAR-10 validation accuracy for $\Sigma\Delta$-Sampler and the full-precision baseline on a 12.98M VGG-8 model with various training configuration and stages.}\label{tab_cl_acc}
\footnotesize
\begin{tabular*}{\textwidth}{@{\extracolsep{\fill}}llccc}
\toprule
Model & Replay & P1 Off (\%) & P2 Off (start $\rightarrow$ end, \%) & P2 On (\%) \\
\midrule
FP        & $20\%$ & $92.83$ & $72.40 \rightarrow 83.85$ & $98.35$ \\
FP        & $50\%$ & $92.99$ & $84.94 \rightarrow 88.23$ & $97.80$ \\
$\Sigma\Delta$-Sampler    & $20\%$ & $92.15$ & $64.84 \rightarrow 77.51$ & $97.60$ \\
$\Sigma\Delta$-Sampler    & $50\%$ & $92.21$ & $78.36 \rightarrow 84.36$ & $97.00$ \\
\bottomrule
\end{tabular*}
\end{table}

The adaptive regularization of discrete parameter update dynamics governed by the $\Sigma\Delta$-Sampler loop naturally lends itself to on-edge adaptation, as new data distributions automatically reshape the loss landscape and alter the network's current state along the annealing trajectory. We evaluate the proposed algorithm in a continual learning setup using CIFAR-10. Specifically, we partition the ten CIFAR-10 classes into 8 offline classes and 2 online classes, training across two distinct phases. Phase 1 (``offline'') exposes the network exclusively to the 8 offline classes. Phase 2 (``online'') introduces the 2 online classes alongside a controlled admixture of offline classes, varied between $20\%$ and $50\%$ to span low- and high-replay regimes. For both phases, we report validation accuracy on the offline and online class groups evaluated separately using the VGG-8 convolutional model. As a baseline, we run an identical protocol on a full-precision counterpart trained with AdamW using the same architecture, learning-rate schedule, and class-mixture sweep. The validation accuracy across setups is shown in Figure~\ref{fig3}f, with detailed statistics compiled in Table~\ref{tab_cl_acc}.

While the state variables that cross the phase boundary differ between the two pipelines, their roles are conceptually analogous. For the $\Sigma\Delta$-Sampler, we checkpoint and reload the complete per-layer accumulator state, comprising the ternary weight tensor $\mathbf{W}$, the $\Sigma\Delta$ integrator buffer $\boldsymbol{\omega}^n$, the dynamic threshold $\mu^n$, the temperature $T^n$, and the step counter $n$. For the full-precision baseline, we checkpoint and reload the latent weights, the AdamW first- and second-moment estimates ($m_1$ and $v$), and the learning-rate schedule state. Consequently, both pipelines enter Phase 2 with a fully populated optimizer state preserved from Phase 1, ensuring neither pipeline benefits from an optimizer reset.

As demonstrated in the results, the $\Sigma\Delta$-Sampler generates continual-learning dynamics qualitatively similar to full-precision AdamW. When incoming data alters the loss landscape established during Phase 1, a subset of stationary parameters previously pinned to local minima are exposed to new directional gradient information, shifting population dynamics back toward an active flip regime. Figure~\ref{fig3}g shows the per-parameter inter-flip interval distributions partitioned by phase. The Phase 1 distributions reproduce the transition from concentrated short intervals to a broad, near-uniform spread observed in the standard benchmark setting. In contrast, the Phase 2 distribution (red) breaks this trend by shifting leftward toward shorter intervals, signifying a population-level reactivation of flip activity rather than continued crystallization. The cumulative distribution inset confirms this shift, showing higher probability mass at short intervals in Phase 2 compared to late Phase 1.

Figure~\ref{fig3}h isolates this effect at the parameter level by plotting the reactivation ratio $R_j = r^{(2)}_j / r^{(1)}_j$ (the ratio of Phase 2 update rate $r^{(2)}_j$ to Phase 1 update rate $r^{(1)}_j$ for parameter $j$) against the Phase 1 update rate $r^{(1)}_j$, overlaid with an inverse polynomial fit. This inverse trend identifies parameters with low Phase 1 activity as the primary reservoir of Phase 2 plasticity which is the individual parameter signature of our contour-hugging mechanism. When the online data mixture alters the local loss surface, parameters that remained dormant in Phase 1 absorb the newly introduced gradient pressure and reactivate, whereas highly active Phase 1 parameters, having already saturated their update budgets, resist parameter reallocation.

\subsection{Scaling $\Sigma\Delta$-Sampler to ImageNet-1000 and language models}\label{sec_res_scaling}

\begin{figure}[h!]
\centering
\includegraphics[width=0.95\textwidth]{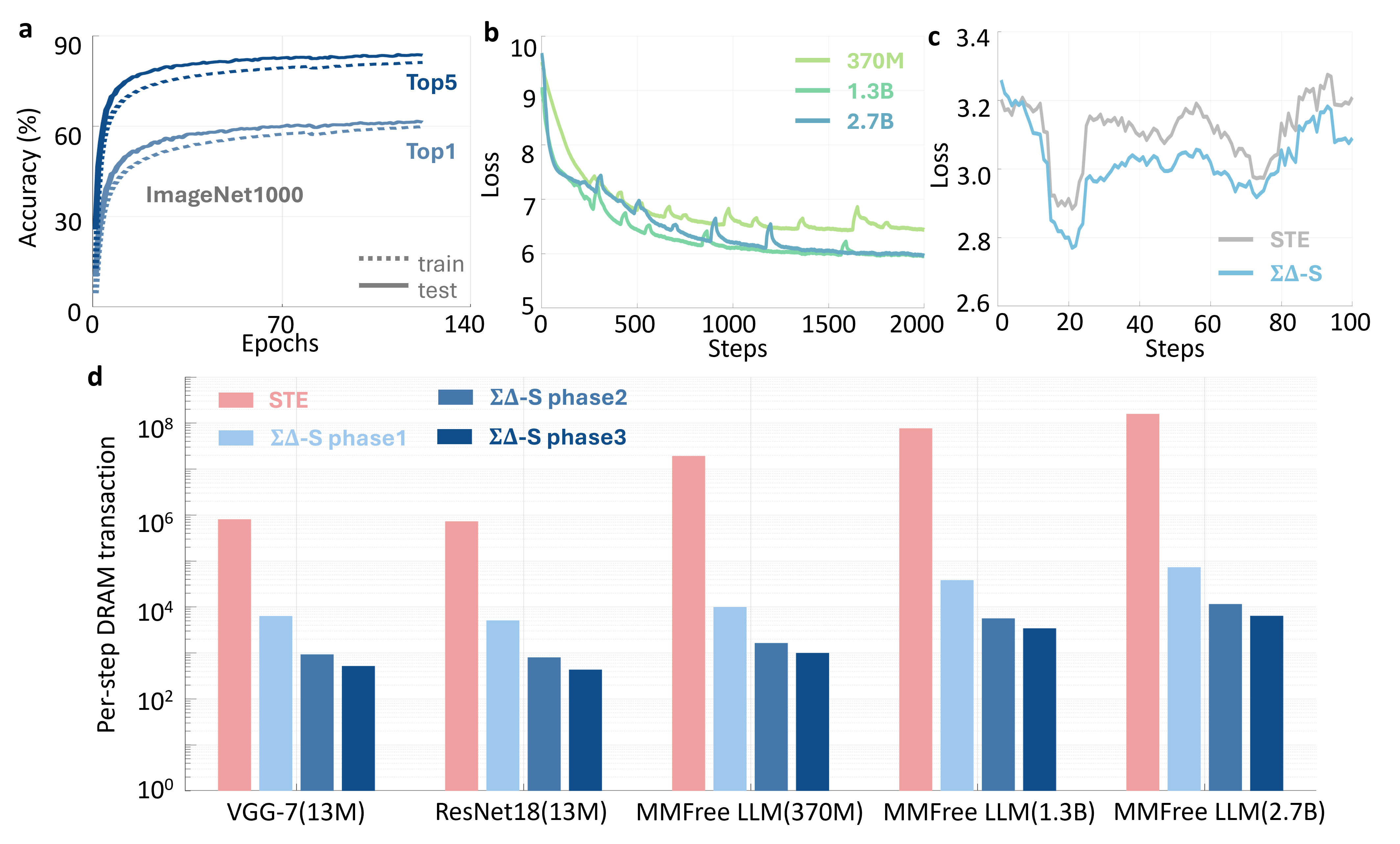}
\caption{\textbf{Scaling up $\Sigma\Delta$-Sampler training on larger dataset (ImageNet1000) and LLMs} (a) Training and test accuracy of ResNet18 on Imagenet1000 with $\Sigma\Delta$-Sampler as optimizer. (b) Training loss of various sized MatMul-Free LLMs on FineWeb-Edu with 532.5K tokens per training step. (c) Continued pre-train on the same dataset from the same quantized checkpoint with STE and $\Sigma\Delta$-Sampler optimizers respectively. (d) Per-step DRAM transactions due to model weight updates for different models, where transactions become sparser for $\Sigma\Delta$-Sampler as training progresses. }\label{fig4}
\end{figure}

Having characterized the $\Sigma\Delta$-Sampler dynamics on CIFAR-10 and under continual adaptation, we test whether the same update rule transfers to substantially more labels, and larger datasets and model families. Figure~\ref{fig4}a reports the training and test Top-1 and Top-5 accuracy trajectories for a ResNet-18 trained end-to-end on ImageNet-1K using the $\Sigma\Delta$-Sampler optimizer. This setup deploys the same parallel-update, noise-shaping, adaptive-thresholding, and blue-noise stochasticity dynamics used at CIFAR-10 scale, requiring no changes to update-rule hyperparameters beyond standard learning-rate rescaling for batch size. The trajectories rise monotonically to a stable plateau with a narrow train--test gap, mirroring the qualitative pattern observed on smaller benchmarks. Table~\ref{tab_imagenet_acc} compares the converged accuracy against ternary-weight and full-precision counterparts on the same model architecture. The $\Sigma\Delta$-Sampler achieves $61.65\%$ Top-1 and $83.81\%$ Top-5 accuracy. This result falls within $0.15\%$ and $0.39\%$ of the reported accruracy from TWN literature~\cite{Liu2023TWN}, which retains a full-precision classification head, whereas all layers in our model are fully discrete. Furthermore, our performance remains within $3.75\%$ and $2.95\%$ of the full-precision reference model for Top-1 and Top-5 accuracy, respectively.

\begin{table}[h]
\centering
\caption{ImageNet-1000 validation accuracy on a ResNet-18 backbone, comparing $\Sigma\Delta$-Sampler against a ternary-weight-network (TWN) baseline with a full-precision classification layer and the full-precision (FP) reference.}\label{tab_imagenet_acc}
\footnotesize
\begin{tabular*}{\textwidth}{@{\extracolsep{\fill}}lcc}
\toprule
Method & Top-5 (\%) & Top-1 (\%) \\
\midrule
FP                                    & $86.76$ & $65.40$ \\
TWN          & $84.20$ & $61.80$ \\
$\Sigma\Delta$-Sampler              & $83.81$ & $61.65$ \\
\bottomrule
\end{tabular*}
\end{table}

To thoroughly evaluate the scalability of the $\Sigma\Delta$-Sampler optimizer, we train MatMul-Free LLMs~\cite{zhu2024scalable} using the proposed method. Detailed descriptions of the model architecture, data preparation, and training configurations are provided in Supplementary Section~\ref{sec_supp_llm}. The architecture matches the original MatMul-Free LLMs, with two key modifications: the BitLinear layers within the MatMul-Free Linear Gated Recurrent Units (MLGRU) are replaced with $\Sigma\Delta$-Linear layers operating in the ternary parameter space, and the language-model head remains continuous, as illustrated in Fig.~\ref{sfig1}. All three models are pre-trained from scratch on a fixed shard of the FineWeb-Edu corpus for $2{,}000$ steps at $532{,}480$ tokens per step, totaling approximately $1.07$ billion tokens. Every run processes identical data in the same order. The $\Sigma\Delta$-Sampler configuration remains identical across all three model sizes. To ensure a stable and smooth descent trajectory in billion-parameter LLM space, we introduce a per-layer, per-step quota on deterministic weight flips, referred to as the update-rate cap. This cap is fixed at $10^{-3}$ across all runs to isolate model scale as the sole experimental variable. The descent dynamics across various update-rate settings are presented in Supplementary Section~\ref{sec_supp_cap}, while the detailed implementation of the cap within the feedback loop is provided in Supplementary Section~\ref{sec_supp_llm_cap}.

Figure~\ref{fig4}b shows the resulting training-loss trajectories, and Table~\ref{tab_llm_loss} reports the converged values. All three model sizes train stably from random initialization along convergent trajectories, establishing that the noise-shaping loop remains well-behaved as the parameter count increases by an order of magnitude and the architecture shifts from convolutional to recurrent/attention-based mechanisms. Scaling the model from $370$\,M to $1.3$\,B parameters yields a clear improvement of $0.486$ on the final 100-step mean loss. The $1.3$\,B and $2.7$\,B models are statistically indistinguishable at this token budget, separated by only $0.016$ against a per-step loss variation of approximately $\pm 0.4$ which is an expected outcome given that both models remain far from data saturation at approximately one billion tokens. We therefore interpret these three points as evidence that the optimizer scales successfully, rather than as a definitive scaling law. Guidance for tuning the update-rate cap across different model scales and training regimes is provided in Supplementary Section~\ref{sec_supp_cap}.

\begin{table}[h]
\centering
\caption{Numerical results of MatMul-Free language-model pre-training on FineWeb-Edu for all sizes. Tail-100 is the mean training loss over the final $100$ steps.}\label{tab_llm_loss}
\footnotesize
\begin{tabular*}{\textwidth}{@{\extracolsep{\fill}}lrccc}
\toprule
Model & Discrete params & Mean loss & Tail-100 loss  \\
\midrule
MMFree-370M & $308.3$ M & $6.8312$ & $6.4544$  \\
MMFree-1.3B & $1.233$ B & $6.3630$ & $5.9684$ \\
MMFree-2.7B & $2.538$ B & $6.4871$ & $5.9846$ \\
\bottomrule
\end{tabular*}
\end{table}

Training from random initialization establishes stability but does not isolate the specific contribution of the update rule. To evaluate STE and $\Sigma\Delta$-Sampler under matched conditions, we conduct a continued pre-training experiment in which both runs start from the same publicly released, quantized checkpoint of a $0.13$\,B ternary MatMul-Free LLM and differ solely in how the ternary weights are updated. Both runs are trained for $100$ steps on the same corpus with identical optimizer settings for continuous parameters. Figure~\ref{fig4}c reports the evaluation loss, with per-run values tabulated in Supplementary Section~\ref{sec_supp_cap_cpt}. It is worth noting that the initial difference in loss value at step $0$ is due to the difference in initializing the scaling factor of both optimizers. While both the $\Sigma\Delta$-Sampler and the STE optimizer follow the same overall loss trajectory across the evaluation window, the $\Sigma\Delta$-Sampler consistently outperforms STE throughout as shown in Figure~\ref{fig4}c. This aligns with the mechanism proposed in Section~\ref{sec_res_benchmarks}, where the adaptive regularization on the discrete parameters enables $\Sigma\Delta$ optimizers to identify the viable decent directions directly in the discrete domain even from a converged checkpoint, whereas the STE update must accumulate sufficient continuous gradient pressure within a latent weight copy to cross a decision boundary, a threshold it largely fails to reach given this learning rate and step count.

\subsection{$\Sigma\Delta$-Sampler hardware advantages}\label{sec_res_hw}

The discrete-throughout-training nature of the $\Sigma\Delta$-Sampler optimizer yields significant hardware benefits in both memory allocation and off-chip memory traffic, making it a highly suitable candidate for training deep learning models on resource-constrained platforms. First, it reduces static memory allocation by eliminating adaptive optimizer states on discrete weights. Second, it reduces dynamic memory traffic generated by weight updates during training, as discrete weights incur write operations only when a flip is committed. Regarding static memory allocation, we restrict our analysis to persistent parameter state—comprising weights, gradients, optimizer moments, and algorithm-specific auxiliary buffers, while excluding intermediate activation memory, which remains identical across optimizers for a given architecture and batch size. As shown in Table~\ref{tab_mem}(a), full-precision training and standard STE quantization-aware training both require four FP quantities per parameter (the weight, the gradient, and the AdamW first and second moments), totaling $16$\,bytes per parameter. Additionally STE packs discrete weights into a single byte, and its overall accounting remains otherwise identical to full precision because it maintains a full-precision latent weight copy in the optimizer. Thus, weight quantization alone provides no net training-time memory savings. 

In contrast, under the $\Sigma\Delta$-Sampler framework, the decision signal $\boldsymbol\omega$ is the only persistent state that needs to be stored across iterations. This state is stored as a pair $W^{\mathrm{on}}, W^{\mathrm{off}} \in \{0,1\}$ packed at one byte per component. Furthermore, in actual implementation the $\Sigma\Delta$ integrator $\boldsymbol{\omega}^n$ reuses the existing gradient buffer attached to the parameter rather than requiring a separate allocation, resulting in a total of $6$\,bytes per discrete parameter. The random draws required by the stochastic branch are consumed directly within the annealing step and require no persistent storage. Crucially, the two AdamW moments (which account for $8$ of the $16$\,bytes in the baseline) are omitted by construction since the annealing update (Eq.~\eqref{eq_method_update_dynamics_full}) relies solely on the current integrator state and a threshold comparison, eliminating the need for historical optimizer states tracking. Continuous parameters, including scale factors, normalization layers, embeddings, and the language-model head, remain under AdamW optimization at $16$\,bytes per parameter but the corresponding memory allocation are negligible compared to the learning parameters.

\begin{table}[h]
\centering
\caption{Persistent training-state accounting. (a) Per-parameter cost of each
stored component. (b) Analytical aggregated memory allocation across models. Activation memory is
excluded throughout, since it is identical across optimizers under matched
architecture and batch size.}\label{tab_mem}
\footnotesize

\begin{tabular*}{\textwidth}{@{\extracolsep{\fill}}lccc}
\toprule
(a) Component & FP / STE & $\Sigma\Delta$-Sampler discrete & $\Sigma\Delta$-Sampler continuous \\
\midrule
Weight or ternary code             & $4$/$1(4)$ & $2$   & $4$ \\
Gradient, or decision signal $\omega^n$ & $4$     & $4$   & $4$ \\
AdamW first moment $m_1$           & $4$     & $0$   & $4$ \\
AdamW second moment $v$            & $4$     & $0$   & $4$ \\
Stochastic draw buffer             & ---     & transient & --- \\
\midrule
\textbf{Total (B/param)}           & $\mathbf{16/17}$ & $\mathbf{6}$ & $\mathbf{16}$ \\
\bottomrule
\end{tabular*}

\bigskip

\begin{tabular*}{\textwidth}{@{\extracolsep{\fill}}lrcccr}
\toprule
(b) Model & Discrete params & Discrete frac. & FP & $\Sigma\Delta$-Sampler & Saving \\
\midrule
VGG-8 (CIFAR-10)     & $12.97$ M & $96.2\%$ & $215.8$ MB & $\mathbf{86.1}$ MB  & $-60.1\%$ \\
ResNet-18 (ImageNet) & $11.68$ M & $89.3\%$ & $209.3$ MB & $\mathbf{92.5}$ MB  & $-55.8\%$ \\
MMFree-370M          & $308.3$ M & $82.4\%$ & $5.99$ GB  & $\mathbf{2.91}$ GB  & $-51.5\%$ \\
MMFree-1.3B          & $1.233$ B & $90.3\%$ & $21.84$ GB & $\mathbf{9.51}$ GB  & $-56.5\%$ \\
MMFree-2.7B          & $2.538$ B & $93.9\%$ & $43.25$ GB & $\mathbf{17.87}$ GB & $-58.7\%$ \\
\bottomrule
\end{tabular*}
\end{table}


Compounding this accounting over each model's parameter inventory produces the aggregate memory footprints reported in Table~\ref{tab_mem}(b). The savings scale directly with the proportion of discrete parameters: $-55.8\%$ on the convolution-dominated ResNet-18 (where per-channel scaling factors carry proportionally higher AdamW overhead) and $-58.7\%$ on the $2.7$\,B language model (where $93.9\%$ of parameters are ternary). At the largest scale, eliminating the two AdamW moments accounts for a $20.30$\,GB reduction, while weight compression contributes an additional $5.08$\,GB saving. Under tight memory constraints, training via the $\Sigma\Delta$-Sampler optimizer can reduce static memory requirements below the capacity threshold of a single accelerator at scales where full-precision training stacks would otherwise OOM (out-of-memory).

Off-chip memory access accounts for a significant fraction of total energy consumption, particularly during training when parameters are iteratively updated~\cite{Horowitz2014ComputeEnergy}. The second major hardware advantage of the $\Sigma\Delta$-Sampler optimizer stems from reducing the off-chip memory traffic generated by weight updates, which directly governs both bandwidth consumption and energy dissipation per step. The $\Sigma\Delta$ integrator undergoes dense read-write operations at every step, but because it reuses the standard gradient buffer, this traffic component is identical to that of full-precision or STE baselines and cancels out. In contrast, weight updates do not cancel: full-precision and STE optimizers apply non-zero updates to every latent weight at each step, forcing a complete rewrite of the weight tensor regardless of update magnitude. Under the $\Sigma\Delta$-Sampler, the persistent parameter is the ternary code itself, which is rewritten only when a discrete flip occurs. Consequently, the weight-update traffic between methods differs by the flip rate, which in the $\Sigma\Delta$-Sampler framework is sparse and decays smoothly alongside the annealing schedule.

We convert measured per-step flip counts into DRAM transaction estimates using a deliberately pessimistic analytical model as detailed in Supplementary Section~\ref{sec_supp_mem}. Off-chip memory is assumed to be accessed at a granularity of 32\,bytes. Thus, a dense rewrite of a full-precision weight tensor requires $N/8$ transactions per step. For the $\Sigma\Delta$-Sampler, we charge a full 32-byte transaction for every flipped weight (granting no multi-flip coalescing credit when adjacent flips share a sector), and we cap the transaction count at $N/32$ when local flip density saturates a sector. Charging a full sector per individual flip represents the worst-case scenario for sparse, spatially uncorrelated updates. The underlying flip counts are recorded empirically at each step during the runs shown in Figure~\ref{fig4}b only the DRAM transaction mapping is analytical.

Figure~\ref{fig4}d reports these findings. The efficiency advantage grows monotonically with model scale, reaching an approximate $1{,}600\times$ transaction reduction on vision models and a $24{,}600\times$ reduction on the $2.7$\,B language model during late-stage training. This reduction is driven by two compounding mechanisms: first, full-precision baseline traffic scales linearly with parameter count while the $\Sigma\Delta$ flip rate does not, as converging parameters undergo crystallization and exhibit increasingly sparse flip activity; second, the adaptive threshold dynamically regulates the global update budget regardless of parameter count, causing late-stage memory traffic to drop far below initial values. Figure~\ref{fig4}d illustrates the average per-step memory transactions, highlighting that full-precision baseline traffic remains constant while $\Sigma\Delta$-Sampler traffic decreases across training phases. Taken together with static footprint savings, the discrete-throughout-training design simultaneously reduces memory footprint and memory movement via a unified structural principle: the parameter being trained is the parameter being stored.

\section{Discussion}\label{sec_discussion}

In this work, we proposed and demonstrated the $\Sigma\Delta$-Sampler, an optimizer that embeds a stochastic sampling-and-annealing step as a noisy quantizer inside a first-order $\Sigma\Delta$ noise-shaping feedback loop. We showed that this mechanism effectively pushes the noise introduced by weight quantization, higher-order frustration, and mini-batch sampling out of the low-frequency adaptation band. As a result, the discrete weight updates follow the smooth descent trajectories embedded within an otherwise rugged loss landscape. This enables the $\Sigma\Delta$-Sampler to be faster and significantly more memory-efficient than standard adaptive-moment optimizers, making it an attractive solution for on-device training and fine-tuning of large binary and ternary models.

Like other Ising solvers (e.g., simulated annealing), the choice of temperature annealing schedule is critical to the optimizer's performance. The temperature schedule used in this work, $T^n = T_0 / (T_0 + n/\tau)$, is asymptotically $1/n$. This places it in the fast-cooling regime motivated by the Cauchy machine of Fast Simulated Annealing (FSA), rather than the classical $1/\log n$ regime of single-spin SA~\cite{SZU1987FSA}. Classical SA requires a $1/\log n$ schedule to guarantee convergence because single-spin proposals only reach a Hamming distance of 1, requiring slower cooling to maintain ergodicity across multi-spin energy barriers~\cite{Chen2025IsingSolver5, Hajek1988OptimalCooling}. FSA relaxes this constraint to a $1/n$ schedule by replacing local proposals with a Cauchy distribution whose infinite variance enables long-range jumps at any temperature, compensating for the faster cooling rate with heavy-tailed exploration. The parallel-flip mechanism in the $\Sigma\Delta$-Sampler produces a per-iteration Hamming distance that is a random variable bounded above by the total parameter count $D$, lacking an infinite tail. Therefore, the structural analogy justifying the $1/n$ schedule in our setting is not a literal Cauchy proposal, but rather the temperature-controlled reach of the parallel-flip mechanism. Just as the Cauchy distribution concentrates as $T \to 0$, the per-iteration Hamming distance contracts in lockstep with temperature, yielding long-range moves at high temperatures and short-range moves at low temperatures without modifying the underlying proposal mechanism. This dynamic property reproduces the dynamic memory access patterns shown in Figure~\ref{fig1}i, leading directly to the dramatic off-chip memory traffic reductions reported in Section~\ref{sec_res_hw}. Thus, the core property justifying fast cooling in FSA is that the proposal is in lockstep with the acceptance temperature, and this attribute is also preserved in our discrete setup, even though the formal heavy-tailed proposal property is not. We treat this connection as a conceptual motivation rather than a rigorous mathematical derivation, as the formal FSA convergence theorem does not directly extend to bounded-support proposals. Instead, we motivate the schedule by analogy and validate it empirically through the planted QUBO experiments in Section~\ref{sec_res_qubo} and the deep neural network benchmarks across Sections~\ref{sec_res_benchmarks} through~\ref{sec_res_scaling}.

Two hardware benefits of $\Sigma\Delta$-Sampler described in Section~\ref{sec_res_hw} is worth emphasizing since the benefits stem from a single underlying mechanism acting on two distinct resources. First, because the discrete weights itself serves as the persistent parameter rather than a projection of a latent copy, there is no full-precision shadow weight to allocate. Second, because the annealing rule relies solely on the current integrator state and a threshold comparison, no adaptive-optimizer moments are maintained for the discrete weights. Consequently, the persistent training state drops to 6\,bytes per ternary parameter compared to 16\,bytes for a full-precision or Straight-Through Estimator (STE) pipeline, a reduction of $55.8\%$ on ResNet-18 and $58.7\%$ on the $2.7$\,B language model, with savings scaling proportionally to the fraction of discrete parameters. STE training captures none of these benefits during optimization because its latent weights are stored as full-precision parameters in the optimizer; thus, weight quantization under STE provides inference-time savings but no training-time footprint reduction. The second benefit concerns bandwidth rather than memory capacity. Full-precision and STE optimizers apply non-zero updates to every weight at every step, forcing a dense rewrite of the entire weight tensor on each iteration. In contrast, under the $\Sigma\Delta$-Sampler, a discrete weight is rewritten only when a flip is committed, and the measured flip rate is both sparse and decreasing over the course of training. Under a deliberately pessimistic transaction model, the resulting reduction in weight-update traffic reaches four orders of magnitude at the billion-parameter scale. Because off-chip memory access dominates the energy budget of training far more than arithmetic operations, this bandwidth reduction represents the primary axis along which the discrete-throughout-training formulation impacts hardware efficiency. This is a benefit that quantization alone cannot deliver, as quantized weights trained through latent copies still incur dense write operations on every step.

We would like to highlight several limitations of the proposed approach which could form the basis of future research directions. First, the bounded-support property of the parallel-flip proposal precludes a direct extension of the FSA convergence theorem to the $1/n$ schedule in the discrete setting. Thus, the asymptotic guarantees provided by classical single-spin SA cannot be applied to the current $\Sigma\Delta$-Sampler algorithm. However, because modern deep learning optimization prioritizes generalization performance rather than finding global energy minima, losing formal discrete convergence bounds is offset by optimization dynamics that favor smooth, generalizable solutions. Empirically, there is still a modest performance gap relative to full-precision baselines, on the order of $1\%$ on CIFAR-10 and $3.75\%$ Top-1 on ImageNet-1K. Furthermore, our language modeling evaluations are conducted on a single pre-training corpus at a sub-saturation token budget, where the primary objective is to demonstrate that the $\Sigma\Delta$-Sampler scales smoothly across dataset and model dimensions with minimal algorithmic modifications. Finally, our memory traffic evaluations rely on an analytical mapping from measured flip counts to hardware DRAM transactions under sector-granularity assumptions, rather than direct hardware performance-counter measurements, and account strictly for weight-update traffic. The most immediate of these limitations can be addressed via straightforward systems engineering using sub-byte packing of the ternary code. This would further compress both static memory footprint and transaction volume, while custom fused kernels (which accumulate gradients directly into the integrator without populating an intermediate buffer) and direct hardware counter profile measurements would validate the empirical traffic model while halving the remaining dense traffic.

The long-term goal for this research is enable training workflows that natively leverage the hardware primitives and software execution pipelines that were originally designed for low-bit inference. Packed low-bit data structures, integer arithmetic units, and optimized execution kernels were engineered to make model deployment lightweight; a training algorithm operating directly on these native representations inherits these efficiencies rather than requiring a parallel full-precision datapath. The present framework advances this paradigm across the full parameter space, however, the activation is not quantized. Also, while our forward pass operates on quantized representations, the backward pass execution relies on continuous gradients because reverse-mode automatic differentiation requires a composition of transposed matrix products. In such composition, quantization errors accumulate multiplicatively with network depth, requiring precise gradients to prevent long-term integrator drift across training steps. There could be two possible approaches to address this bottleneck. The first approach pushes quantization deeper into the backward path, trading gradient precision for a fully low-precision execution pipeline. The second approach could use backpropagation-free gradient estimators. Alternative gradient estimators, such as Direct Feedback Alignment (DFA)~\cite{Arild2016DFA} (which replaces transposed weight matrices in the backward pass with fixed random projections) and Simultaneous Perturbation Stochastic Approximation (SPSA)~\cite{Spall1998spsa} (which estimates descent directions via localized forward perturbations), bypass the precision-sensitive chained matrix multiplications that make standard backpropagation sensitive to quantization noise. Under either approach, error accumulation across deep multiplicative chains is eliminated, significantly relaxing activation precision requirements. Note that the sampler/quantizer in $\Sigma\Delta$-Sampler processes gradients differently from the traditional gradient-descent approaches, as it never scales a gradient by a learning rate to perform an additive weight update. Instead, it processes a directional decision signal (the integrator) evaluated against an adaptive threshold; the noise-shaping loop (Section~\ref{sec_methods_sigma_delta}) is specifically designed to extract a reliable, low-frequency descent vector from noisy per-iteration inputs. Thus, the update is engineered to tolerate stochastic noise in its driving signal which provides a substantially more robust host for low-fidelity gradient estimators than algorithms where gradient magnitudes directly dictate update step sizes. If both weights and activations were made fully discrete throughout training, the overall execution loop would reduce to low-precision forward passes, single-accumulator per-parameter tracking, and sparse discrete state transitions, fundamentally bridging the gap between training and inference architectures. At this point of convergence, the core update primitives transition from software optimizations to native hardware opportunities: binary state representations, stochastic comparisons, and temporal integration map directly to physical operations in in-memory computing and neuromorphic crossbar architectures, where the sampler/quantizer and $\Sigma\Delta$ integrator can be physically realized via device-level stochastic comparators and analog memory elements rather than emulated in digital logic.

\section{Methods}\label{sec3_methods}

\subsection{Exact Multilinear Decomposition of the Discrete Loss}\label{sec_methods_walsh}

Let $D$ index every discrete weight in the network, so a configuration is $\mathbf{W}\in\{-1,+1\}^D$ and the loss is a real-valued function on $2^D$ points, fully specified by the values it takes there. We derive its exact polynomial form constructively. For a target $a_i\in\{-1,+1\}$ the quantity $(1+a_iW_i)/2$ equals one when $W_i=a_i$ and zero otherwise, since $a_iW_i=\pm 1$. Multiplying over coordinates gives the indicator of a single corner,
\begin{equation}
    P_{\mathbf{a}}(\mathbf{W}) \;=\; \prod_{i=1}^{D}\frac{1+a_iW_i}{2},
    \label{eq_methods_indicator}
\end{equation}
which vanishes whenever any coordinate mismatches. Exactly one indicator is non-zero at any $\mathbf{W}$, so
\begin{equation}
    \mathcal{L}(\mathbf{W}) \;=\; \sum_{\mathbf{a}\in\{-1,+1\}^D} \mathcal{L}(\mathbf{a})\,P_{\mathbf{a}}(\mathbf{W}),
    \label{eq_methods_interp}
\end{equation}
which is the retrieval of a tabulated value written as a polynomial in $\mathbf{W}$.

Expanding~\eqref{eq_methods_indicator}, each bracket contributes either $1$ or $a_iW_i$, and a choice of brackets is a subset $S\subseteq\{1,\ldots,D\}$, giving $2^D$ terms
\begin{equation}
    \prod_{i=1}^{D}\bigl(1+a_iW_i\bigr)
    \;=\; \sum_{S} \Bigl(\prod_{i\in S} a_i\Bigr)\Bigl(\prod_{i\in S} W_i\Bigr).
    \label{eq_methods_expand}
\end{equation}
Substituting into~\eqref{eq_methods_interp}, exchanging the two finite sums, and extracting $\prod_{i\in S}W_i$ from the inner sum yields
\begin{equation}
    \mathcal{L}(\mathbf{W}) \;=\; \sum_{m=0}^{D}\ \sum_{|S|=m} J^{(m)}_S \prod_{i\in S} W_i ,
    \qquad
    J^{(m)}_S \;=\; \mathbb{E}_{\mathbf{W}}\!\left[\mathcal{L}(\mathbf{W})\prod_{i\in S} W_i\right],
    \label{eq_methods_walsh}
\end{equation}
the expectation being the uniform average over corners.

The representation is exact and model-agnostic. It is unique by counting, since $2^D$ subsets supply $2^D$ coefficients against $2^D$ interpolation conditions, and the monomials are squarefree because $W_i^2=1$ collapses repeated indices. Equation~\eqref{eq_methods_walsh} is the Walsh--Fourier expansion of a Boolean function~\cite{ODonnell2014}. Comparing it with Eq.~\eqref{eq:ising}, the discrete weights are spins, a subset $S$ is an interaction clause, and $J^{(m)}_S$ is the $m$-spin coupling, with $m=|S|$ counting the weights participating irreducibly in one term. Order zero is a constant irrelevant to optimization, order one acts as external fields, and a non-vanishing coupling of order $m$ is a failure of the loss to decompose into contributions of fewer than $m$ weights acting independently.

\subsection{Discrete Neural Networks as Energy-Based Systems}\label{sec_methods_problem}

In the Ising model, a binary spin vector $\mathbf{s} \in \{-1,+1\}^D$ minimizes a quadratic energy functional
\begin{equation}
    H(\mathbf{s}) \;=\; \tfrac{1}{2}\,\mathbf{s}^\intercal\,\mathbf{Q}\,\mathbf{s},
\end{equation}
where $\mathbf{Q}$ is a symmetric coupling matrix; because $s_i^2 = 1$, diagonal terms do not affect optimization and $Q_{ii}$ may be set to zero without loss of generality. This discrete formulation underlies Hopfield networks, Boltzmann machines, and a long line of physics-inspired learning systems, in which inference and learning are interpreted as descent over a finite configuration space of size $2^D$.~\cite{barahona1982Ising1, Hopfield1982, Hopfield1984}

Modern deep neural networks predominantly operate over continuous parameter domains $\mathbf{W} \in \mathbb{R}^D$, optimized by gradient descent on differentiable objectives. Binary neural networks (BNNs) instead constrain weights to
\begin{equation}
    \mathbf{W} \in \{-1,+1\}^D,
\end{equation}
restoring a combinatorial configuration space analogous to spin systems. As established in the introduction in Eq.~\eqref{eq:ising}--\eqref{eq:bnn-ising}, the inference function of a BNN decomposes into a hierarchy of $m$-th order spin interactions $J^{(m)}$ given by Eq.~\eqref{eq_methods_walsh}. Training the discrete neural networks is therefore
\begin{equation}
    \min_{\mathbf{W}\in\{-1,+1\}^D}\, \mathcal{L}(\mathbf{W}),
    \label{eq_methods_bnn_objective}
\end{equation}
where $\mathcal{L}$ is an empirical loss evaluated under data sampling. Although $\mathcal{L}$ is implicitly defined by nonlinear layer compositions rather than written out as an explicit polynomial Hamiltonian, it remains an energy functional over a finite discrete state space of cardinality $2^D$.


Throughout this section we develop the framework in the binary case $\mathbf{W}\in\{-1,+1\}^D$ for clarity. The same framework extends without modification to ternary weights $\mathbf{W}\in\{-1,0,+1\}^D$ via the decomposition 
$W_{i} = W^{\mathrm{on}}_{i} - W^{\mathrm{off}}_{i}$ with $W^{\mathrm{on}}_{i},\, W^{\mathrm{off}}_{i}\in\{1,0\}$, of which the decompisition is detailed in Section~\ref{sec_methods_ternary}. The sampling dynamics, $\Sigma\Delta$ feedback, and adaptive thresholding of Section~\ref{sec_methods_sigma_delta} through Section~\ref{sec_method_gradacc_aptthres} apply to each component independently.

\subsection{Noise-shaping in $\Sigma\Delta$-Sampler Extracts the Smooth Contour within the Discrete Loss Landscape}\label{sec_methods_sigma_delta}

The Boltzmann-weighted parallel update in single-spin SA, applied directly to the per-batch STE gradient $g^n_{i}$, sees the rugged optimization contour as explained through the multi-order Ising decomposition in Eq.~\eqref{eq:bnn-ising}. What the $\Sigma\Delta$ sampling optimizer should follow is not this noisy per-iteration surface but the smooth low-frequency contour underlying it, the slowly-varying envelope contributed by the dominant low-order terms of the high-order Ising expansion of Eq.~\eqref{eq:bnn-ising}. To recover that contour from the noisy per-iteration stream, we choose the decision signal $\mathbf{\omega}_i^n$ through first-order $\Sigma\Delta$ noise-shaping loop~\cite{Chakrabartty2013NoiseShaped}. The feedback loop structurally shapes the noise contribution out of the low-frequency adaptation band where the smooth descent signal lives and into the high-frequency Nyquist neighborhood where they are rejected by the thresholding and sampling process rather than allowed to corrupt the parameter trajectory.

For each coordinate $i$ we maintain an integrator state $\omega^n_{i}$ initialized at $\omega^0_{i} = 0$ and a discrete update signal $d^n_{i}\in\{-2 W^{n-1}_{i},\,0\}$ that records the realized weight increment at iteration $n$. The $\Sigma\Delta$-Sampler update cycle is
\begin{align}
    \omega^n_{i} &= \omega^{n-1}_{i} + g^n_{i} - \Delta\beta^n_{i}\, d^{n-1}_{i}, \quad\text{(integrator)} \label{eq_methods_sd_int}\\
    d^n_{i} &\sim \mathcal{P}\!\left(\omega^n_{i},\, T^n\right), \quad\text{(sampler)} \label{eq_methods_sd_quant}\\
    W^n_{i} &= W^{n-1}_{i} + d^n_{i}, \quad\text{(weight update)} \label{eq_methods_sd_weight}
\end{align}
where $\mathcal{P}(\,\cdot\,,\,T)$ denotes the sampler acting on $\omega^n_{i}$ in place of $g^n_{i}$ as the decision signal. The feedback gain ${\Delta\beta}^n_{i}={T^n}^2({\beta}^n_{i}-{\beta}^{n-1}_{i})$ is attenuated by the same temperature schedule used in the quantizer to ensure the dynamic range of the feedback signal matches the decision signal $\omega_{i}^n$. Here, ${\beta}^n_{i}$ is a sub-coordinate learnable auxiliary parameter, which is a standard practice that rescales the discrete parameters in the forward inference~\cite{Liu2023TWN, zhu2024scalable}. In the derivation, the model weights are denoted as a $D$-dimensional vector to remain model agnostic. In actual implementation, for a linear layer all discrete parameter on the same row share the same gain scalar and for convolutional layers all parameters in the same kernel share the same gain. 
To quantify the noise-shaping effect, we adopt the standard additive-noise model in which the per-iteration quantization noise is summarized into a single additive term $\mathbf{q}^n$ that is approximately white across the iteration band as shown below for a certain parameter,
\begin{equation}
    {W}_{i}^n \;=\; {W}_{i}^{n-1} - \eta^n\left({g}^{n-1*}_{i} - q_{i}^n\right).
    \label{eq_methods_noisyGD}
\end{equation}
Here ${g}^{*}_{i}$ denotes an ideal noiseless gradient signal derived from full-precision parameter space that differs from the noisy STE gradient ${g}^{}_{i}$ obtained in the discrete parameter space. $q_{i}$ captures the entirety of quantization noise in the proposed $\Sigma\Delta$ training loop and $\eta^n$ represents the aggregate parameter update rate modulation effect of the $\Sigma\Delta$ feedback. To analyze the frequency response of noise-shaping on the learning parameters, we linearize the quantized noisy gradient descent in Eq.~\eqref{eq_methods_noisyGD} around a local minimum $\mathbf{W}^*$ with local Hessian $\mathbf{H}$, treating ${\eta}^n$ as approximately constant on the relevant timescale. For simplicity we show the $z$-domain output of the closed-loop $\Sigma\Delta$ recursion of Eqs.~\eqref{eq_methods_sd_int}--\eqref{eq_methods_sd_weight} for a scalar parameter
\begin{equation}
    \mathbf{E}(z) \;=\; -\Bigl[\mathbf{I} + z^{-1}\bigl(\eta\,\mathbf{H} - \mathbf{I}\bigr)\Bigr]^{-1}\,\eta\,(1 - z^{-1})\,\mathbf{Q}(z),
    \label{eq_methods_ntf}
\end{equation}
where ${E}(z)$ denotes the spectrum of the error on the parameter $\mathbf{e} = \mathbf{W} - \mathbf{W^*}$, and ${Q}(z)$ is the frequency domain signal of quantization error $\mathbf{q}$. The detailed derivation is presented in Supplementary Section~\ref{sec_supp_sd_sd}. It could be observed that the noise transfer function $\mathrm{NTF}(z) = 1 - z^{-1}$ acts as a first-order high-pass on the additive quantization noise. The high-frequency content of the additive noise is structurally pushed into the near-Nyquist neighborhood while the low-frequency adaptation band is preserved. Compared to an unshaped open-loop quantized-gradient-descent baseline that we show in Supplementary Section~\ref{sec_supp_sd_qgd}, whose error spectrum lacks the $(1 - z^{-1})$ shaping factor, the closed-loop mean-square error within an adaptation band of bandwidth $\gamma_a$ relative to the iteration rate $\gamma_s$ is reduced by approximately
\begin{equation}
    \frac{\mathcal{E}_{\Sigma\Delta}}{\mathcal{E}_{\mathrm{QGD}}} \;\approx\; \frac{\gamma_a^2}{3\,\gamma_s^2}, \qquad \gamma_a \ll \gamma_s,
    \label{eq_methods_mse_ratio}
\end{equation}
so the in-band noise floor scales as $(\gamma_a/\gamma_s)^2$ relative to the unshaped baseline. The narrower the adaptation band relative to the iteration rate, the larger the noise-shaping gain, and the closer the $\Sigma\Delta$-Sampler's per-iteration decision stream is driven by the smooth low-frequency descent signal rather than the wideband per-iteration noise. The full derivation under the small-signal linearization, including the per-component transfer functions and the closed-loop pole expression, is given in Supplementary Section~\ref{sec_supp_sd}.

\subsection{{Ternary Weights via Decomposition}}\label{sec_methods_ternary}

The framework extends from binary to ternary weights without altering the loop of Section~\ref{sec_methods_sigma_delta}. We represent each ternary weight as the difference of two binary components,
\begin{equation}
    W_{i} \;=\; W^{\mathrm{on}}_{i} - W^{\mathrm{off}}_{i},
    \qquad W^{\mathrm{on}}_{i},\, W^{\mathrm{off}}_{i} \in \{0,1\},
    \qquad W_{i} \in \{-1,0,+1\},
    \label{eq_methods_ternary_decomp}
\end{equation}
so that each component is a binary variable to which the flip rule of Eq.~\eqref{eq_method_update_dynamics_full} applies unchanged, with the clamp acting on $\{0,1\}$ rather than $\{-1,+1\}$. Because $\partial W_{i}/\partial W^{\mathrm{on}}_{i} = +1$ and $\partial W_{i}/\partial W^{\mathrm{off}}_{i} = -1$, the two components see antipodal decision signals. We exploit this to retain a single integrator per weight rather than one per component, driving $W^{\mathrm{on}}$ with $+\omega^n_{i}$ and $W^{\mathrm{off}}$ with $-\omega^n_{i}$,
\begin{equation}
    d^{\mathrm{on},n}_{i} = \mathcal{P}\!\left(+\omega^n_{i},\, T^n\right),
    \qquad
    d^{\mathrm{off},n}_{i} = \mathcal{P}\!\left(-\omega^n_{i},\, T^n\right).
    \label{eq_methods_ternary_signals}
\end{equation}
The two components therefore share the accumulator, the gradient, the temperature $T^n$, the log-domain scale $\bar{G}^n$, the adaptive threshold $\mu^n$, and the AGC state of Section~\ref{sec_method_gradacc_aptthres}. The only quantity not shared is the random draw, which is sampled independently for each component so that their stochastic branches are not perfectly correlated. Since $W^{\mathrm{off}}$ carries no gradient of its own under this scheme, it is excluded from the autograd graph and no second gradient buffer is allocated for it.

The feedback path closes on the effective weight rather than on either component. The forward pass depends only on $W_{i}$, so the quantization error that the loop is designed to cancel is the error in the effective weight, and the integrator recursion of Eq.~\eqref{eq_methods_sd_int} becomes
\begin{equation}
    \omega^n_{i} \;=\; \omega^{n-1}_{i} + g^n_{i}
    \;-\; \tfrac{1}{2}\,\Delta\beta^n_{i}\,
    \bigl(W^n_{i} - W^{n-1}_{i}\bigr).
    \label{eq_methods_ternary_feedback}
\end{equation}

\subsection{Adaptive Thresholding via Log-Domain Gradient Statistics}\label{sec_method_gradacc_aptthres}

The $\Sigma\Delta$ feedback of Section~\ref{sec_methods_sigma_delta} produces an integrator state $\omega^n_{i}$ whose magnitudes span several orders of magnitude across layers and training stages. We partition coordinates at each iteration into a deterministic descent set and a stochastic exploration set,
\begin{equation}
    \mathcal{D}^n = \{i\,:\,|\omega^n_{i}|>\mu^n\}, \qquad
    \mathcal{S}^n = \{i\,:\,|\omega^n_{i}|\le\mu^n\},
    \label{eq_methods_partition}
\end{equation}
and set the threshold $\mu^n$ by a feedback loop that locks the realized fraction $|\mathcal{D}^n|/D$ to a prescribed schedule.

A robust per-iteration scale is obtained from an EMA over the log-magnitude $\log(|\omega^n_{i}|+\varepsilon)$, giving a geometric-mean estimate $\bar{G}^n$ insensitive to the heavy-tailed magnitude distribution. An automatic gain control (AGC) loop then drives $\mu^n$ toward a target above-threshold fraction that ramps harmonically from a stochastic-dominated initial value to a deterministic-dominated steady-state value,
\begin{equation}
    \rho_t^n = \rho_{\mathrm{init}} + \bigl(\rho_{\mathrm{final}} - \rho_{\mathrm{init}}\bigr)\,\frac{n}{n+\tau_{\mathrm{thld}}},
    \label{eq_methods_target_ramp}
\end{equation}
with defaults $\rho_{\mathrm{init}}=0.2$ and $\rho_{\mathrm{final}}=0.7$. Letting $\rho^n_{\mathrm{ema}}$ be the EMA of the realized above-threshold fraction, the AGC gain and resulting threshold are
\begin{equation}
    \mathrm{gain}^n = \bigl(\rho^n_{\mathrm{ema}}/\rho_t^n\bigr)^{\alpha_{\mathrm{agc}}}, \qquad
    \mu^n = \mu_{\min} + \gamma\,\bar{G}^n\,\mathrm{gain}^n.
    \label{eq_methods_mu}
\end{equation}
When too many coordinates pass the threshold the gain rises and $\mu^n$ is pushed up; when too few pass it the gain falls and $\mu^n$ is pulled down. The early-stochastic to late-deterministic balance characteristic of crystallization therefore emerges from the AGC ramp without any manual schedule, with $\rho_{\mathrm{init}}=0.2$ corresponding to $80\%$ of coordinates in the stochastic regime at the start of training and $\rho_{\mathrm{final}}=0.7$ corresponding to $30\%$ at convergence.

The combined update rule is
\begin{equation}
    W^n_{i} =
    \begin{cases}
        -W^{n-1}_{i}, & \text{if } |\omega^n_{i}|>\mu^n \text{ and } \mathrm{sign}(\omega^n_{i})=W^{n-1}_{i}, \\[4pt]
        -W^{n-1}_{i}, & \text{w.p. } \exp\!\bigl(-|\omega^n_{i}|/\bar{G}^nT^n\bigr) \text{ if } |\omega^n_{i}|\le\mu^n, \\[4pt]
        \;\;\, W^{n-1}_{i}, & \text{otherwise}.
    \end{cases}
    \label{eq_method_update_dynamics_full}
\end{equation}

\subsection{Model Architectures and $\Sigma\Delta$-Layer Placement}\label{sec_methods_models}

All experiments use two $\Sigma\Delta$-Sampler layer types, \texttt{$\Sigma\Delta$\_Conv2D} and \texttt{$\Sigma\Delta$\_Linear}, as drop-in replacements for their standard convolutional and fully-connected counterparts. Each layer carries a discrete weight tensor updated by the annealing dynamics of Section~\ref{sec_method_gradacc_aptthres} together with a small population of continuous parameters that remain under AdamW, namely a per-output-row scale (\texttt{row\_scaling}) and bias for linear layers and a per-output-channel/input-channel multiplier (\texttt{kernel\_multiplier}) for convolutional layers. Normalization layers, activation functions, and dropout are unchanged from the standard architectures. In contrast to the common practice of exempting the first and last layers from quantization, every learnable weight layer in our vision models is discrete, including the input-facing convolution and the classifier head.

For CIFAR-10 we use two backbones of the VGG family. VGG-7 consists of six discrete convolutional layers of widths $128, 128, 256, 256, 512, 512$, all $3{\times}3$ and each followed by batch normalization, ReLU, and spatial dropout, with $2{\times}2$ max-pooling after each width pair, followed by a single \texttt{$\Sigma\Delta$\_Linear} classifier, giving approximately $5$M discrete parameters. VGG-8 extends the same convolutional stack with a two-layer classifier, \texttt{$\Sigma\Delta$\_Linear}$(8192{\to}1024)$, batch normalization, and \texttt{$\Sigma\Delta$\_Linear}$(1024{\to}10)$, giving eight discrete weight layers and $12.97$M discrete parameters. Unless otherwise stated the CIFAR-10 dynamics reported in Section~\ref{sec_res_benchmarks} are measured on VGG-8. For ImageNet-1000 we use a ResNet-18 that follows the standard torchvision topology, with a discrete $7{\times}7$ stem convolution, four stages of two basic blocks each at widths $64, 128, 256, 512$ in which both $3{\times}3$ block convolutions and the $1{\times}1$ downsampling shortcut convolutions are discrete layers, batch normalization throughout, and an \texttt{$\Sigma\Delta$\_Linear}$(512{\to}1000)$ classifier, with stem, block-level, and pre-classifier dropout for regularization. The vision models train in fp32 and use ternary weights $W\in\{-1,0,+1\}$ under the $W^{\mathrm{on}}-W^{\mathrm{off}}$ decomposition of Section~\ref{sec_methods_ternary}, with the binary case $W\in\{-1,+1\}$ recovered as a special case and used only where explicitly indicated.

For language modeling we integrate the $\Sigma\Delta$-Sampler machinery into the MatMul-Free transformer of~\cite{zhu2024scalable}, implemented as \texttt{MMFreeBitTransformerForCausalLM}. Each block applies a pre-norm RMSNorm followed by multi-head attention with rotary position embeddings, then a second RMSNorm followed by a SwiGLU MLP whose gate projection produces the gate and value halves jointly. All six linear projections per block, the attention projections $q$, $k$, $v$, $o$ and the MLP projections \texttt{gate\_proj} and \texttt{down\_proj}, are \texttt{BitLinear} layers that combine an input RMSNorm with 8-bit per-token activation quantization, and every one of them is converted to a discrete layer operating in ternary mode, in which the weight code is stored as the pair $W^{\mathrm{on}}, W^{\mathrm{off}} \in \{0,1\}$ with effective weight $W = W^{\mathrm{on}} - W^{\mathrm{off}} \in \{-1,0,+1\}$ and a per-output-row continuous scale in AdamW. The token embedding, the language-model head, the block-level RMSNorm scales, and the per-\texttt{BitLinear} input norms remain continuous. Parameters are held in fp32 with bf16 applied by autocast, so the continuous parameters carry full fp32 AdamW state. We train three scales of this architecture, summarized in Table~\ref{tab:llm_configs}, in which the discrete fraction of the parameter inventory grows from $82.4\%$ to $93.9\%$ as the model widens while the embedding and normalization cost stays approximately constant. Full training configuration, data preparation, and the definition of the update-rate cap are given in Supplementary Section~\ref{sec_supp_llm}.

\begin{table}[h]
\centering
\caption{MatMul-Free LLM configurations}\label{tab:llm_configs}
\scriptsize
\setlength{\tabcolsep}{4pt}
\begin{tabular}{@{}lccccrr@{}}
\toprule
Model & $H$ & Layers & Heads & Intermediate & Discrete & Fraction \\
\midrule
MMFree-370M & 1024 & 24 & 16 & 2816 & 308.3 M & 82.4\% \\
MMFree-1.3B & 2048 & 24 & 32 & 5632 & 1.23 B  & 90.3\% \\
MMFree-2.7B & 2560 & 32 & 40 & 6912 & 2.54 B  & 93.9\% \\
\bottomrule
\end{tabular}
\end{table}

In all settings the discrete weights are excluded from the AdamW optimizer entirely. Each training iteration performs the standard forward and backward pass, applies the AdamW step to the continuous parameters only, and then invokes the anneal step of Eq.~\eqref{eq_method_update_dynamics_full} on every discrete layer, using the integrator state $\omega^n_{i}$ carried in the gradient buffer as the decision signal.

\subsection{Visualization: Principal Component Analysis of Update Trajectories}\label{sec_methods_pca}

To visualize the evolution of the parameter-space dynamics of the three optimizers on the planted QUBO of Section~\ref{sec_res_qubo}, we apply principal component analysis (PCA) to their discrete update streams, following a dimensionality-reduction procedure similar to that used for neural population dynamics in the literature~\cite{friedrich2001dynamic, stopfer2003intensity, gangopadhyay2021gtnn}. For each variant $v\in\{\mathrm{SF},\,\mathrm{PF\text{-}20},\,\Sigma\Delta\}$ we record the discrete-update vector $\mathbf{d}^n_v\in\{-2,0,+2\}^d$ at every iteration $n=1,\ldots,N$ and stack the resulting sequence into a per-variant trajectory matrix $\mathbf{D}_v\in\mathbb{R}^{N\times d}$. The 3 matrices are concatenated row-wise into a joint matrix $\mathbf{D}_{\mathrm{joint}}\in\mathbb{R}^{3N\times d}$, mean-centered along the coordinate axis, and decomposed into principal components via the eigendecomposition of its empirical covariance. The top three principal vectors define a shared three-dimensional embedding into which each variant's time-ordered update sequence is projected, producing the 3D trajectories shown in Fig.~\ref{fig2}c. Constructing the PCA basis jointly across variants ensures that the axes are common to all 3 and that trajectory comparisons are directly meaningful. The inset panel d isolates the \text{SF}, \text{PF-20}, and $\Sigma\Delta$ variants on a tighter scale so that the structural characteristics of the $\Sigma\Delta$ trajectories can be read off without the visual dominance of the diverging \text{PF-20} curves.

\section*{Data Availability}\label{sec_data}
The planted-QUBO convergence traces, BNN training metrics, and benchmark evaluation data generated in this study are available at \url{https://github.com/aimlab-wustl/SD-SA}.

\section*{Code Availability}\label{sec_code}
The $\Sigma\Delta$-Sampler training framework, including the Python implementation of the $\Sigma\Delta$ simulated annealing optimizer, the blue-noise rank-mask construction for convolutional layers, and all benchmark training scripts, are available at \url{https://github.com/aimlab-wustl/SD-SA}.

\section*{Acknowledgments}\label{sec_ack}
 This work was supported by the National Science Foundation grants CNS - FuSE2 - 2425444, ECCS - 2332166 and FET -2551604. Claude and Gemini AI tools were used for correcting typos, structuring sentences, to proof-check some of the mathematical models and analysis presented in the supplementary information. The Authors acknowledge the National Artificial Intelligence Research Resource (NAIRR) Pilot and Purdue Rosen Center for Advanced Computing (RCAC) Anvil compute cluster for contributing to this research result. We also acknowledge feedback from the reviewers of the pre-print version of this manuscript for suggesting corrections and pointing out typos.

\section*{Author Contributions}\label{sec_contributions}
Z.C. and Z.X. formulated the BNN/TNN training algorithm based on classical simulated annealing; S.C and Z.C. proposed and formulated the $\Sigma\Delta$-sampler for training QUBO, BNN and TNNs. Z.C. and Z.X. ran the benchmark vision experiments and Y.W provided assistance in setting up the vision benchmarks and datasets. Z.C. optimized the $\Sigma\Delta$-Sampler parameters for training. R-J.Z and J.E helped in setting up the MatMult-free LLM benchmarks and Z.C., R-J.Z, and Z.X. ran the $\Sigma\Delta$-Sampler on these benchmarks. All authors/co-authors contributed to proof-reading and writing of the manuscript.  

\section*{Competing interests}\label{sec_COI}
The authors declare no competing interests.

\newpage
\setcounter{section}{0}
\setcounter{equation}{0}
\setcounter{figure}{0}
\setcounter{table}{0}
\renewcommand{\thesection}{S\arabic{section}}
\renewcommand{\thefigure}{S\arabic{figure}}
\renewcommand{\thetable}{S\arabic{table}}
\renewcommand{\theequation}{S\arabic{equation}}
\makeatletter
\renewcommand\theHsection{supp.\thesection}
\renewcommand\theHfigure{supp.\thefigure}
\renewcommand\theHtable{supp.\thetable}
\renewcommand\theHequation{supp.\theequation}
\makeatother

\section*{Supplementary Information}\label{sec_supp}

\section{Noise-Shaping Analysis for Discrete-Space $\Sigma\Delta$-Sampler Training}\label{sec_supp_sd}

The $\Sigma\Delta$ noise-shaping described in the main text relies on an additive-noise model of the per-iteration $\Sigma\Delta$-Sampler updates. Under this model, the spectral behavior of the $\Sigma\Delta$ loop reduces to the spectral behavior of a first-order $\Sigma\Delta$ gradient-descent recursion of the kind studied in~\cite{Chakrabartty2013NoiseShaped}, which provides closed-form expressions for adaptation-band noise reduction relative to an unshaped baseline. In this section, we adapt that analysis to the $\Sigma\Delta$-Sampler setting. The derivation proceeds in three steps: (i) lumping the three $\Sigma\Delta$-Sampler noise sources into a single additive noise term so that the standard $\Sigma\Delta$ framework applies, (ii) deriving the error transfer function around a local minimum, and (iii) integrating the resulting power spectrum across the adaptation band to obtain a closed-form expression for the noise-shaping gain.

\subsection{Additive-noise model of the per-iteration $\Sigma\Delta$-Sampler update}\label{sec_supp_sd_noise}

Around a local minimum $\mathbf{W}^*$ of the discrete loss landscape, we approximate the per-iteration parameter trajectory as a small-signal continuous variable. This is the standard small-signal linearisation used for spectral analysis of discrete-amplitude adaptation systems~\cite{Chakrabartty2013NoiseShaped}. Let $\mathbf{e}^n = \mathbf{W}^n - \mathbf{W}^*$ denote the error vector at iteration $n$ and let
\begin{equation}
    \mathbf{g}^n \;=\; \nabla_{\mathbf{W}}\mathcal{L}(\mathbf{W}^*)\,+\,\mathbf{H}\,\mathbf{e}^{n-1} \;+\; \mathbf{q}^n
    \label{eq_supp_sd_grad_decomp}
\end{equation}
denote the per-iteration STE gradient, decomposed into its Taylor expansion around $\mathbf{W}^*$ (with $\mathbf{H} = \nabla^2_{\mathbf{W}}\mathcal{L}(\mathbf{W}^*)$ the Hessian) and an additive noise term $\mathbf{q}^n$. The leading term $\nabla\mathcal{L}(\mathbf{W}^*)$ vanishes at the local minimum, leaving the linearised model
\begin{equation}
    \mathbf{g}^n \;\approx\; \mathbf{H}\,\mathbf{e}^{n-1} \;+\; \mathbf{q}^n.
    \label{eq_supp_sd_grad_lin}
\end{equation}

The noise term $\mathbf{q}^n$ captures three structurally distinct contributions identified in the main text. First, mini-batch sampling noise arises from the finite-sample estimate of $\nabla\mathcal{L}$ and is approximately white on timescales longer than a single batch step. Second, straight-through estimator (STE) quantization error is introduced by treating the binary projection as the identity in the backward pass; this error is also approximately white because the per-iteration discrepancy is uncorrelated with the slowly varying mean gradient. Third, single-bit sampler stochasticity results from the Boltzmann acceptance rule, which replaces a continuous update with a $\{-2W^{n-1},0\}$-valued discrete update. Within the small-signal model, all three contributions act as additive perturbations to the deterministic linearized gradient and are aggregated into $\mathbf{q}^n$ with a cross-power spectral density that we approximate as flat across the iteration band. This reflects the canonical white-noise assumption used in $\Sigma\Delta$ converters~\cite{Chakrabartty2013NoiseShaped}, which has been shown empirically to hold for binary quantizers under additive dither~\cite{Chakrabartty2013NoiseShaped}. Under this assumption, the per-component power spectral density of $\mathbf{q}^n$ is
\begin{equation}
    |Q(\gamma)|^2 \;=\; \sigma_q^2/\gamma_s
    \label{eq_supp_sd_q_psd}
\end{equation}
within the iteration band $[-\gamma_s,\gamma_s]$, with $\sigma_q^2$ the per-coordinate noise variance and $\gamma_s$ the iteration rate.

\subsection{Unshaped baseline: quantized raw gradient descent}\label{sec_supp_sd_qgd}

The baseline (without noise-shaping) against which we benchmark the $\Sigma\Delta$ loop is the quantized-gradient-descent (QGD) recursion in which the sampler acts directly on $\mathbf{g}^n$ without any feedback path, so that the realised parameter increment is the noisy gradient itself. Under the small-signal model, this is equivalent to the recursion
\begin{equation}
    \mathbf{W}^n \;=\; \mathbf{W}^{n-1} \;-\; \eta\,\bigl[\mathbf{g}^n\bigr],
    \label{eq_supp_sd_qgd}
\end{equation}
where $\eta$ is the effective small-signal step size set jointly by the temperature $T^n$, the dynamic threshold $\mu^n$, and the per-coordinate scaling $\beta_{i}$ of the $\Sigma\Delta$-Sampler update rule in the main text. Substituting the linearised gradient~\eqref{eq_supp_sd_grad_lin} into~\eqref{eq_supp_sd_qgd} and rearranging in terms of the error $\mathbf{e}^n$,
\begin{equation}
    \mathbf{e}^n \;=\; \mathbf{e}^{n-1} \;-\; \eta\,\bigl[\mathbf{H}\,\mathbf{e}^{n-1} + \mathbf{q}^n\bigr].
    \label{eq_supp_sd_qgd_err}
\end{equation}

Taking the $z$-transform of both sides and solving for the error spectrum gives
\begin{equation}
    \mathbf{E}(z) \;=\; -\Bigl[\mathbf{I} + z^{-1}\bigl(\eta\,\mathbf{H} - \mathbf{I}\bigr)\Bigr]^{-1}\,\eta\,\mathbf{Q}(z).
    \label{eq_supp_sd_qgd_spec}
\end{equation}
For analytic tractability we restrict the analysis to a scalar parameter (or equivalently to a single eigenvector of $\mathbf{H}$ with eigenvalue $H \in \mathbb{R}_{>0}$). The error transfer function becomes
\begin{equation}
    E(z) \;=\; \frac{-\eta\,Q(z)}{1 - z^{-1} + \eta\,H\,z^{-1}}.
    \label{eq_supp_sd_qgd_scalar}
\end{equation}
At frequencies $\gamma \ll \gamma_s$, $z^{-1}\approx 1 - j\gamma/\gamma_s$ and the magnitude transfer function is
\begin{equation}
    \frac{|E(\gamma)|}{|Q(\gamma)|} \;=\; \frac{1}{H}\,\left[1 + \frac{\gamma^2}{\gamma_p^2}\right]^{-1/2},
    \label{eq_supp_sd_qgd_tf}
\end{equation}
where the closed-loop pole is
\begin{equation}
    \gamma_p \;=\; \frac{\eta\,\gamma_s\,H}{1 - \eta\,H}.
    \label{eq_supp_sd_pole}
\end{equation}
Under a brick-wall post-processing filter $P(\gamma) = \mathbb{1}[|\gamma|\le\gamma_a]$ that retains only the adaptation band, the mean-square error of the QGD error spectrum, by Parseval's theorem, is
\begin{equation}
    \mathcal{E}_{\mathrm{QGD}} \;=\; \int_{-\gamma_a}^{\gamma_a} |E(\gamma)|^2\,d\gamma
    \;=\; \frac{2\sigma_q^2}{H^2\,\gamma_s}\,\gamma_p\,\tan^{-1}\!\left(\frac{\gamma_a}{\gamma_p}\right).
    \label{eq_supp_sd_qgd_mse}
\end{equation}
Equation~\eqref{eq_supp_sd_qgd_mse} is the $\Sigma\Delta$-Sampler analogue of the QGD adaptation-band MSE of~\cite{Chakrabartty2013NoiseShaped}. It exposes the fundamental cost of using a per-iteration noisy gradient as the sampler input. The MSE at the local minimum is set by the noise variance $\sigma_q^2$ at the iteration rate and is reduced only by widening the adaptation pole or narrowing the post-filter, both of which trade off against tracking bandwidth.

\subsection{First-order $\Sigma\Delta$ loop}\label{sec_supp_sd_sd}

The $\Sigma\Delta$ noise shaping loop replaces the open-loop noisy increment of~\eqref{eq_supp_sd_qgd} with a feedback recursion in which the unrealised portion of the previous update is added back into the integrator state and re-presented to the quantizer at the next step. Under the additive-noise model~\eqref{eq_supp_sd_grad_lin}, this recursion is equivalent to driving the parameter trajectory with the differenced noise $\mathbf{q}^n - \mathbf{q}^{n-1}$ instead of the raw $\mathbf{q}^n$, so that
\begin{equation}
    \mathbf{W}^n \;=\; \mathbf{W}^{n-1} \;-\; \eta\,\bigl[\mathbf{H}\,\mathbf{e}^{n-1} + \mathbf{q}^n - \mathbf{q}^{n-1}\bigr].
    \label{eq_supp_sd_sd_rec}
\end{equation}
This is the small-signal interpretation of the loop equations~\eqref{eq_methods_sd_int}--\eqref{eq_methods_sd_weight} of the Methods under the assumption that the feedback gain $\Delta\beta^n_{i} = (T^n)^2\bigl(\beta^n_{i} - \beta^{n-1}_{i}\bigr)$ is approximately constant on the relevant timescale and that the quantizer output $d^n$ can be expressed as $g^n + (1-z^{-1})q^n$ in the $z$-domain. Taking the $z$-transform of~\eqref{eq_supp_sd_sd_rec}, the error spectrum is
\begin{equation}
    \mathbf{E}(z) \;=\; -\Bigl[\mathbf{I} + z^{-1}\bigl(\eta\,\mathbf{H} - \mathbf{I}\bigr)\Bigr]^{-1}\,\eta\,(1 - z^{-1})\,\mathbf{Q}(z),
    \label{eq_supp_sd_sd_spec}
\end{equation}
which differs from the QGD result~\eqref{eq_supp_sd_qgd_spec} by the additional $(1 - z^{-1})$ noise-shaping factor on $\mathbf{Q}(z)$. In the scalar case the magnitude transfer function becomes
\begin{equation}
    \frac{|E(\gamma)|}{|Q(\gamma)|} \;=\; \frac{\gamma}{H\,\gamma_s}\,\left[1 + \frac{\gamma^2}{\gamma_p^2}\right]^{-1/2}.
    \label{eq_supp_sd_sd_tf}
\end{equation}
The closed-loop pole $\gamma_p$ is the same as for the QGD case; what changes is the appearance of $\gamma/\gamma_s$ in the numerator, which expresses the first-order high-pass shaping of the additive noise. Low-frequency components of $\mathbf{q}^n$ are attenuated by a factor proportional to $\gamma/\gamma_s$ relative to the QGD spectrum~\eqref{eq_supp_sd_qgd_tf}, while high-frequency components are preserved.

\subsection{Adaptation-band MSE and the noise-shaping gain}\label{sec_supp_sd_mse}

Integrating the squared magnitude transfer function~\eqref{eq_supp_sd_sd_tf} across the adaptation band gives the mean-square error of the $\Sigma\Delta$-Sampler recursion as
\begin{equation}
    \mathcal{E}_{\Sigma\Delta} \;=\; \int_{-\gamma_a}^{\gamma_a}\!\frac{\gamma^2 |Q(\gamma)|^2}{H^2\,\gamma_s^2}\,\left[1 + \frac{\gamma^2}{\gamma_p^2}\right]^{-1}\!d\gamma
    \;=\; \frac{2\sigma_q^2 \gamma_p^2}{H^2\,\gamma_s^3}\,\,\left[\frac{\gamma_a}{\gamma_p} - \tan^{-1}\!\left(\frac{\gamma_a}{\gamma_p}\right)\right].
    \label{eq_supp_sd_sd_mse}
\end{equation}
The ratio of the noise-shaped and unshaped MSEs is
\begin{equation}
    \frac{\mathcal{E}_{\Sigma\Delta}}{\mathcal{E}_{\mathrm{QGD}}} \;=\; \,\frac{\gamma_p^2}{\gamma_s^2}\,\frac{\dfrac{\gamma_a}{\gamma_p} - \tan^{-1}\!\left(\dfrac{\gamma_a}{\gamma_p}\right)}{\tan^{-1}\!\left(\dfrac{\gamma_a}{\gamma_p}\right)}.
    \label{eq_supp_sd_ratio}
\end{equation}
For $\gamma_a \ll \gamma_p$, expanding $\tan^{-1}(\gamma_a/\gamma_p) \approx \gamma_a/\gamma_p - \tfrac{1}{3}(\gamma_a/\gamma_p)^3$ gives
\begin{equation}
    \frac{\mathcal{E}_{\Sigma\Delta}}{\mathcal{E}_{\mathrm{QGD}}} \;\approx\; \frac{1}{3}\,\frac{\gamma_p^2}{\gamma_s^2}\,\frac{\gamma_a^2}{\gamma_p^2}\;=\;\frac{\gamma_a^2}{3\,\gamma_s^2},
    \label{eq_supp_sd_ratio_lowf}
\end{equation}
so the closed-loop $\Sigma\Delta$ MSE in the adaptation band is smaller than the open-loop QGD MSE by a factor that scales as $(\gamma_a/\gamma_s)^2$. The narrower the adaptation band relative to the iteration rate, the larger the noise-shaping gain, which is independent of $H$, $\eta$, and $\gamma_p$. In the $\Sigma\Delta$-Sampler setting, the adaptation band $\gamma_a$ is the bandwidth of the smooth low-frequency loss-landscape contour that the $\Sigma\Delta$ loop is designed to expose, and $\gamma_s$ is the iteration rate of the sampler.

The dual interpretation of this result, described in the main text and confirmed empirically by the steeper-than-$1/f^2$ spectra of Fig.~\ref{fig3}d,e, is that the same first-order $(1-z^{-1})$ factor that high-pass-shapes the noise in the parameter spectrum also high-pass-shapes the high-frequency content of the integrator state $\omega^n$. The two views are equivalent under Eq.~\eqref{eq_supp_sd_sd_spec}, since both reflect the unity-signal, $(1-z^{-1})$-noise transfer function of the first-order loop. In the limit $T^n \to 0$ the deterministic-flip threshold $\mu^n$ contracts faster than the stochastic-flip rate as shown in Eq.~\eqref{eq_methods_mu}, and the effective step size $\eta$ contracts accordingly, so the closed-loop pole $\gamma_p$ contracts and the adaptation band $\gamma_a$ over which Eq.~\eqref{eq_supp_sd_ratio_lowf} holds is automatically squeezed toward the low-frequency descent component of the gradient. The annealing schedule and the noise-shaping mechanism therefore act in concert, with annealing concentrating the adaptation band on the smooth descent signal and noise shaping attenuating the additive noise within that band.

\section{Algorithmic Implementation of $\Sigma\Delta$-Sampler}\label{sec_supp_algo}

\subsection{Per-iteration anneal-step pseudocode}\label{sec_supp_algo_pseudo}

The per-iteration $\Sigma\Delta$-Sampler update implemented in \texttt{$\Sigma\Delta$\_Layer.anneal\_step()} in the released code can be written compactly as Algorithm~\ref{alg_supp_anneal_step}. Inputs are the current binary weight $\mathbf{W}^{n-1}$, the accumulated gradient buffer $\mathbf{g}^n$ which already encodes the $\Sigma\Delta$ integrator $\omega^n$ via cross-mini-batch accumulation, and the per-layer scaling $\beta$. Outputs are the updated binary weight $\mathbf{W}^n$ and the updated gradient buffer for use as $\omega^{n+1}$ in the next call.
\begin{algorithm}[h!]
\caption{Per-iteration $\Sigma\Delta$-Sampler anneal step (binary case).}\label{alg_supp_anneal_step}
\begin{algorithmic}[1]
\State \textbf{State:} $\mathbf{W}^{n-1}\in\{-1,+1\}^D$; $\mathbf{g}^n\in\mathbb{R}^D$ (gradient buffer = $\omega^n$); EMA $m^{n-1}$, $\rho^{n-1}_{\mathrm{ema}}$; step $n$.
\State \textbf{Params:} $T_0,\tau,\beta_{\mathrm{ema}},\gamma,\mu_{\min}$, $\rho_{\mathrm{init}},\rho_{\mathrm{final}},\tau_{\mathrm{thld}},\alpha_{\mathrm{agc}},\beta_{\mathrm{agc}}$.
\State $T^n \leftarrow T_0 / (T_0 + n/\tau)$ \Comment{Hyperbolic cooling}
\State $m^n \leftarrow \beta_{\mathrm{ema}}\,m^{n-1} + (1-\beta_{\mathrm{ema}})\,\tfrac{1}{D}\sum_{i}\log(|g^n_{i}|+\varepsilon)$; $\bar{G}^n\leftarrow\exp(m^n)$
\State $\rho_t^n \leftarrow \rho_{\mathrm{init}} + (\rho_{\mathrm{final}}-\rho_{\mathrm{init}})\,n/(n+\tau_{\mathrm{thld}})$ \Comment{AGC target ramp}
\State $\mathrm{gain}^n \leftarrow (\rho^{n-1}_{\mathrm{ema}}/\rho_t^n)^{\alpha_{\mathrm{agc}}}$
\State $\mu^n \leftarrow \mu_{\min} + \gamma\,\bar{G}^n\,\mathrm{gain}^n$ \Comment{Dynamic threshold}
\For{each coordinate $i$ \textbf{in parallel}}
    \If{$\omega^n_{i} < -\mu^n$}
        \State $\mathbf{W}^n_{i} \leftarrow +1$ \Comment{Deterministic descent push to $+1$}
    \ElsIf{$\omega^n_{i} > +\mu^n$}
        \State $\mathbf{W}^n_{i} \leftarrow -1$ \Comment{Deterministic descent push to $-1$}
    \Else
        \State Draw $u_{i}$ (white noise for FC, blue-noise rank for conv; cf.~\S\ref{sec_supp_algo_bn})
        \If{$u_{i} < \exp\!\bigl(-|\omega^n_{i}|/(\bar{G}^n\,T^n)\bigr)$}
            \State $\mathbf{W}^n_{i} \leftarrow -\mathbf{W}^{n-1}_{i}$ \Comment{Thermal flip}
        \Else
            \State $\mathbf{W}^n_{i} \leftarrow \mathbf{W}^{n-1}_{i}$
        \EndIf
    \EndIf
\EndFor
\State $\mathbf{\omega}^n \leftarrow \mathbf{\omega}^n + \mathbf{g}^n - \,\Delta\beta\,(\mathbf{W}^n - \mathbf{W}^{n-1})$ \Comment{$\Sigma\Delta$ feedback into gradient buffer}
\State $\rho^n_{\mathrm{ema}} \leftarrow \beta_{\mathrm{agc}}\,\rho^{n-1}_{\mathrm{ema}} + (1-\beta_{\mathrm{agc}})\,\tfrac{1}{D}\sum_{i}\mathbb{1}[|g^n_{i}|>\mu^n]$
\Ensure $\mathbf{W}^n\in\{-1,+1\}^D$, updated $(\mathbf{g}^n, m^n, \rho^n_{\mathrm{ema}})$
\end{algorithmic}
\end{algorithm}

The ternary extension is obtained by running Algorithm~\ref{alg_supp_anneal_step} independently on each of the two component buffers $\mathbf{W}^{\mathrm{on}}\in\{0,1\}^D$ and $\mathbf{W}^{\mathrm{off}}\in\{0,1\}^D$ that compose $\mathbf{W} = \mathbf{W}^{\mathrm{on}} - \mathbf{W}^{\mathrm{off}}$, with $\mathbf{W}^{\mathrm{off}}$ driven by gradient $-\mathbf{g}^n$ (produced automatically by autograd because $\mathbf{W}^{\mathrm{off}}$ enters the effective weight with a negative coefficient).

\subsection{Blue-noise rank-mask construction}\label{sec_supp_algo_bn}

The blue-noise rank mask used for stochastic flip sampling on convolutional layers is precomputed once at initialisation per kernel size $k$ by greedy farthest-point sampling on the $k\times k$ grid. Algorithm~\ref{alg_supp_bn} reproduces the construction as implemented in \texttt{build\_blue\_noise\_rank\_mask()}.
\begin{algorithm}[h!]
\caption{Blue-noise rank-mask construction on a $k\times k$ grid.}\label{alg_supp_bn}
\begin{algorithmic}[1]
\State $\mathcal{P}\leftarrow\{(r,c):0\le r,c < k\}$; $N\leftarrow k^2$
\State Select $p_0$ uniformly at random from $\mathcal{P}$
\State $\mathcal{S}\leftarrow\{p_0\}$; $\mathrm{order}\leftarrow[p_0]$
\For{$i = 1,\ldots,N-1$}
    \State $p_i \leftarrow \arg\max_{p\in\mathcal{P}\setminus\mathcal{S}}\,\min_{s\in\mathcal{S}}\,\|p-s\|^2$
    \State $\mathcal{S}\leftarrow\mathcal{S}\cup\{p_i\}$; append $p_i$ to $\mathrm{order}$
\EndFor
\For{$i = 0,\ldots,N-1$}
    \State Let $(r_i,c_i)=\mathrm{order}_i$; assign rank $R^{\mathrm{base}}[r_i, c_i] \leftarrow i/N\in[0,1)$
\EndFor
\Ensure rank tensor $R^{\mathrm{base}}\in[0,1)^{k\times k}$
\end{algorithmic}
\end{algorithm}

At each anneal step, the per-channel-pair blue-noise variate $U_{c_{\mathrm{out}},c_{\mathrm{in}},u,v}\in[0,1)$ used in Algorithm~\ref{alg_supp_anneal_step} (line 11) is generated from $R^{\mathrm{base}}$ by drawing a uniform rotation $r\in\{0,90^\circ,180^\circ,270^\circ\}$ and a Bernoulli horizontal flip applied to $R^{\mathrm{base}}$, then adding a per-channel-pair phase shift $\phi_{c_{\mathrm{out}},c_{\mathrm{in}}}\sim\mathrm{Unif}[0,1)$ and a per-position dither $\eta_{c_{\mathrm{out}},c_{\mathrm{in}},u,v}\sim\mathrm{Unif}[0,1/N)$, both modulo 1.

\subsection{Hyperparameter defaults}\label{sec_supp_algo_hparams}

Table~\ref{tab_supp_hparams} lists the default values of the $\Sigma\Delta$-Sampler-specific hyperparameters used throughout the experiments of Sections~\ref{sec_res_qubo} through~\ref{sec_res_continual_adaptation}. These are the values defined in the \texttt{Model\_HParams} dataclass of the released code and are the only knobs of the algorithm beyond the standard learning rate, weight decay, and batch size of the AdamW optimizer that handles the continuous scaling parameters.
\begin{table}[h]
\centering
\caption{Default hyperparameters of the $\Sigma\Delta$-Sampler per-layer dynamics. Identifiers in monospace are the field names of the \texttt{Model\_HParams} dataclass.}\label{tab_supp_hparams}
\footnotesize
\begin{tabular*}{\textwidth}{@{\extracolsep{\fill}}llll}
\toprule
Symbol & Identifier & Default & Role \\
\midrule
$T_0$                       & \texttt{temp\_init}                 & 2.0   & Initial temperature \\
$\tau$                      & \texttt{temp\_decay}                & 5.0   & Cooling time-constant \\
$\beta_{\mathrm{ema}}$      & \texttt{ema\_beta\_log}             & 0.95  & EMA decay for log-gradient statistic \\
$\mu_{\min}$                & \texttt{thld\_min}                  & 0.0   & Threshold floor \\
$\gamma$                    & \texttt{thld\_gain}                 & 1.0   & Threshold gain on $\bar{G}^n$ \\
$\tau_{\mathrm{thld}}$      & \texttt{thld\_tau}                  & 2000  & AGC ramp time-constant (steps) \\
$\rho_{\mathrm{init}}$      & \texttt{thld\_target\_pruned\_init} & 0.2   & AGC initial deterministic-fraction target \\
$\rho_{\mathrm{final}}$     & \texttt{thld\_target\_pruned}       & 0.7   & AGC steady-state deterministic-fraction target \\
$\alpha_{\mathrm{agc}}$     & \texttt{thld\_agc\_alpha}           & 0.3   & AGC gain exponent \\
$\beta_{\mathrm{agc}}$      & \texttt{thld\_agc\_beta}            & 0.99  & EMA decay for pruned-fraction tracker \\
---                         & \texttt{sigma\_delta}               & true  & Enables $\Sigma\Delta$ feedback \\
---                         & \texttt{sigma\_delta\_temp\_scale}  & true  & Scales feedback by current $T^n$ \\
---                         & \texttt{use\_dynamic\_threshold}    & ``hard'' & Threshold-gated deterministic/stochastic split \\
---                         & \texttt{ternary}                    & false & If true, $\mathbf{W}\in\{-1,0,+1\}^D$ via $\mathbf{W}^{\mathrm{on}}-\mathbf{W}^{\mathrm{off}}$ \\
\bottomrule
\end{tabular*}
\end{table}

\section{Class-incremental continual-learning protocol}\label{sec_supp_cl}

We complement the description in Section~2.4 of the main text with the additional protocol details required for reproducibility. The ten CIFAR-10 classes are partitioned into 8 offline classes and 2 online classes, with the partition fixed across all replay-fraction conditions. The same VGG-8 architecture used for the standard-benchmark experiments of the main text is trained from scratch in two phases. Phase 1 runs for 150 epochs on the offline class group, using the full $\Sigma\Delta$-Sampler training pipeline of the the Methods with the hyperparameters of Table~\ref{tab_supp_hparams}. At the end of phase 1, the per-layer decision signal state, comprising the ternary weight tensor $\mathbf{W}$, the gradient buffer that encodes the $\Sigma\Delta$ integrator, the log-gradient EMA $m^n$, the pruned-fraction EMA $\rho^n_{\mathrm{ema}}$, the dynamic threshold $\mu^n$, and the step counter $n$, is checkpointed and reloaded as the starting point of phase 2. Phase 2 runs for 50 additional epochs on a data stream comprising the 2 online classes and a controlled admixture of the 8 offline classes, with the admixture fraction varied between 20\% and 50\% to span low- and high-replay conditions. Each minibatch is drawn from the mixed stream so that the gradient signal seen by the sampler at each anneal step combines new and replayed examples in the specified proportion.

The full-precision baseline runs the identical class-incremental protocol on a continuous-weight VGG-8 of the same architecture trained with AdamW, using the same learning-rate schedule, batch size, and class-admixture sweep. The FP baseline differs from $\Sigma\Delta$-Sampler only in the optimizer (AdamW vs. the annealing dynamics of Algorithm~\ref{alg_supp_anneal_step}) and in the parameter precision (fp32 vs. ternary). No replay buffer, regularizer, or task-aware gating is applied to either model. The reactivation analysis of Fig.~\ref{fig3}h uses normalised per-parameter update rates computed over the final epoch of each phase, with the rate of a parameter defined as the number of accepted flips divided by the number of anneal steps in the epoch.

\section{Language-model experimental setup}\label{sec_supp_llm}

This section documents the architecture, data pipeline, and optimization settings for the language-model experiments of Section~\ref{sec_res_scaling}. All quantities are as run.

\subsection{Model family and architecture}\label{sec_supp_llm_arch}

\begin{figure}[h!]
\centering
\includegraphics[width=0.95\textwidth]{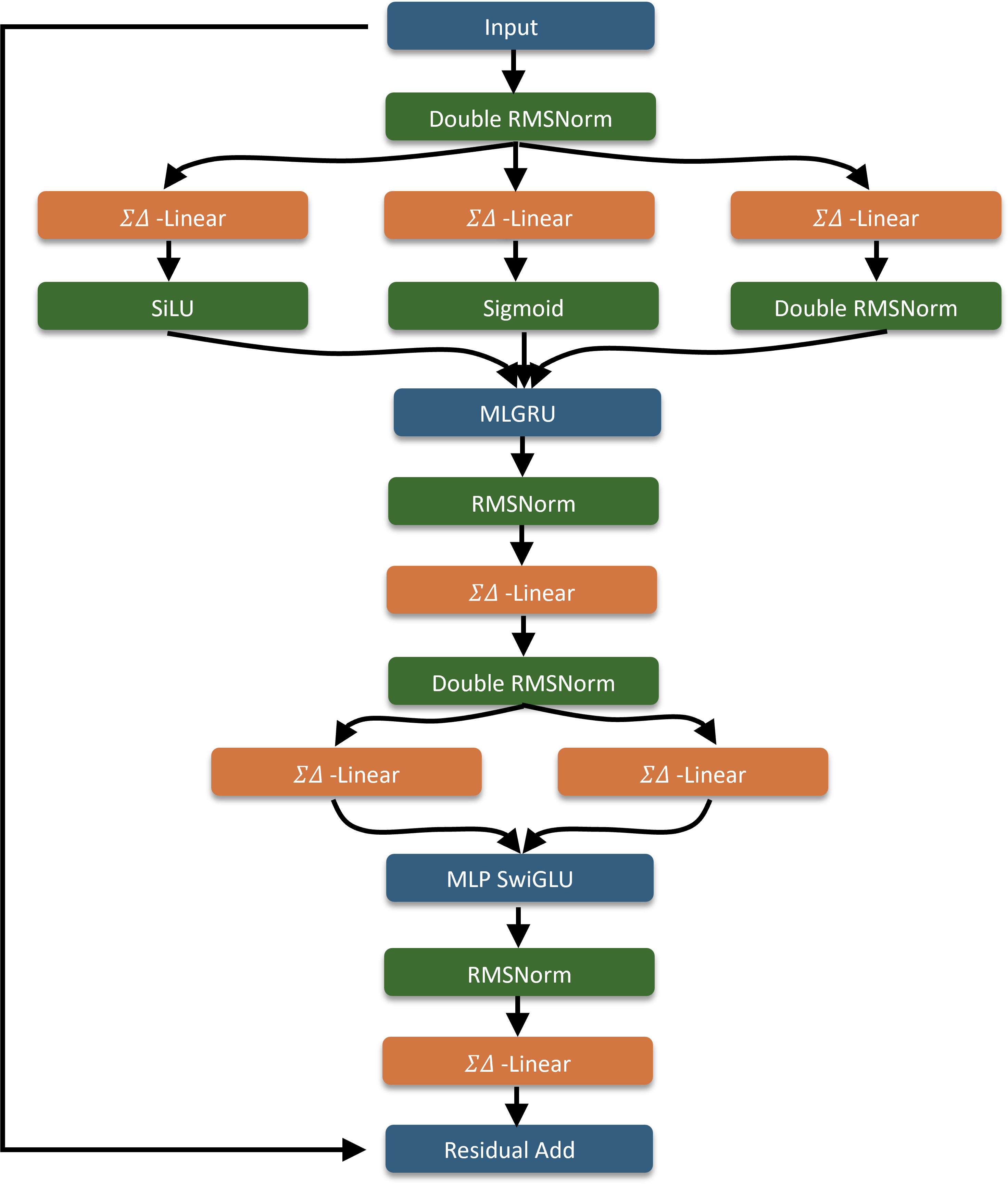}
\caption{\textbf{Computational graph of a single MatMul-free language model layer.} }\label{sfig1}
\end{figure}

The language models are MatMul-Free LLMs~\cite{zhu2024scalable}, where the \texttt{BitLinear} layers in the original architecture is replaced in-place with the \texttt{$\Sigma\Delta$-Linear} layers as shown in Fig.~\ref{sfig1}. Each decoder block applies a pre-norm RMSNorm followed by multi-head attention with rotary position embeddings, then a second RMSNorm followed by a SwiGLU feed-forward network whose gate projection emits the gate and value halves jointly. Every linear projection inside a block is a \texttt{FusedBitLinear} layer, which fuses an input RMSNorm with per-token 8-bit activation quantization into a single kernel and applies W1.58 ternary weight quantization. Six projections per block are placed under $\Sigma\Delta$-Sampler, namely \texttt{attn.q\_proj}, \texttt{attn.k\_proj}, \texttt{attn.v\_proj}, \texttt{attn.o\_proj}, \texttt{mlp.gate\_proj}, and \texttt{mlp.down\_proj}. The language-model head is excluded and remains continuous, as are the token embedding, the block-level RMSNorm scales, and the per-\texttt{BitLinear} input norms. Table~\ref{tab_supp_llm_arch} lists the configurations.

\begin{table}[h]
\centering
\caption{Language-model configurations. Vocabulary is $32{,}000$ for all sizes, embedding and language-model head are untied, and the SwiGLU intermediate width is $256\lceil (8H/3)/256\rceil$ for hidden width $H$. \emph{Discrete params} counts the ternary parameters under $\Sigma\Delta$-Sampler and \emph{layers} counts the number of $\Sigma\Delta$-Sampler layers.}\label{tab_supp_llm_arch}
\footnotesize
\begin{tabular*}{\textwidth}{@{\extracolsep{\fill}}lccccrr}
\toprule
Model & $H$ & Layers & Heads & Intermediate & Discrete params & $\Sigma\Delta$ layers \\
\midrule
MMFree-370M & $1024$ & $24$ & $16$ & $2816$ & $308{,}281{,}344$ & $144$ \\
MMFree-1.3B & $2048$ & $24$ & $32$ & $5632$ & $1{,}233{,}125{,}376$ & $144$ \\
MMFree-2.7B & $2560$ & $32$ & $40$ & $6912$ & $2{,}537{,}553{,}920$ & $192$ \\
\midrule
12L768 (S5.3) & $768$ & $12$ & $12$ & $2048$ & $109{,}510{,}656$ & $72$ \\
\bottomrule
\end{tabular*}
\end{table}

\subsection{Data}\label{sec_supp_llm_data}

All runs use a single pinned shard of the FineWeb-Edu corpus, subset \texttt{sample/10BT}, containing $182{,}101$ documents. Text is tokenized with the \texttt{MMfreeLM-370M} Llama tokenizer at a vocabulary of $32{,}000$, streamed, and packed into blocks of $1024$ tokens. The token geometry is $520 \times 1024 = 532{,}480$ tokens per optimizer step, and pre-training runs for $2{,}000$ steps, giving approximately $1.065$ billion tokens presented. Because this exceeds the size of the single shard, the stream wraps and the models see a partial second epoch; the figure should therefore be read as tokens presented rather than unique tokens. All runs consume identical data in identical order, so cross-configuration comparisons are unaffected.

\subsection{Optimization of the continuous parameters}\label{sec_supp_llm_opt}

Continuous parameters are optimized with AdamW at learning rate $2\times10^{-5}$, $\beta = (0.9, 0.95)$, $\varepsilon = 10^{-8}$, weight decay $0.1$, in the fused implementation, under a cosine decay to zero over the run with a single warmup step. Global gradient-norm clipping is applied at $1.0$ over the continuous parameters only, since the discrete weights carry an integrator state rather than a step direction and including them would set the clip factor from parameters that the clip is not intended to control. Parameters are constructed in fp32 and bf16 is applied by autocast. Runs use a single NVIDIA H100 80\,GB device with no sharding. The random seed is $42$.

We note one implementation detail because it materially affects reproduction. When model parameters are constructed directly in bf16 rather than fp32, AdamW updates smaller than the bf16 representable increment round to zero, and over $20$ steps only $29.7\%$ of \texttt{BitLinear} weight elements change value, against $100\%$ under fp32 parameters. The effect is severe for a straight-through baseline, whose weights move only through accumulated continuous updates, and largely absent for $\Sigma\Delta$-Sampler, whose ternary weights move by discrete flips. Any comparison between the two under bf16 parameters is therefore biased in favour of $\Sigma\Delta$-Sampler. All results reported in this work use fp32 parameters.

\subsection{$\Sigma\Delta$-Sampler configuration}\label{sec_supp_llm_sa}

The $\Sigma\Delta$-Sampler settings are identical across the three pre-training sizes; only the four architecture keys of Table~\ref{tab_supp_llm_arch} differ. Table~\ref{tab_supp_llm_sa} lists them.

\begin{table}[h]
\centering
\caption{$\Sigma\Delta$-Sampler configuration for the language-model experiments, identical at all three pre-training scales.}\label{tab_supp_llm_sa}
\footnotesize
\begin{tabular*}{\textwidth}{@{\extracolsep{\fill}}lll}
\toprule
Setting & Value & Meaning \\
\midrule
mode                       & direct ternary & $W = W^{\mathrm{on}} - W^{\mathrm{off}}$, components in $\{0,1\}$ \\
$T_0$                      & $2.0$   & initial temperature \\
$\tau$                     & $0.5$   & $T^n = T_0/(T_0 + n/\tau)$ \\
$\Sigma\Delta$ feedback    & enabled & second-order temperature scaling, $(T^n)^2$ \\
single accumulator         & enabled & difference mode $(\omega^{\mathrm{on}} - \omega^{\mathrm{off}})/2$ \\
pure-integration warmup    & $20$ steps & gradient integrated before feedback engages \\
$\beta_{\mathrm{ema}}$     & $0.95$  & log-magnitude EMA decay \\
$\mu_{\min}$, $\gamma$     & $0.0$, $1.0$ & threshold floor and gain \\
$\tau_{\mathrm{thld}}$     & $10{,}000$ & AGC target ramp constant \\
$\rho_{\mathrm{init}} \to \rho_{\mathrm{final}}$ & $0.2 \to 0.7$ & AGC target above-threshold fraction \\
$\alpha_{\mathrm{agc}}$, $\beta_{\mathrm{agc}}$ & $0.3$, $0.99$ & AGC exponent and EMA decay \\
update frequency           & every step & anneal step invoked once per optimizer step \\
\bottomrule
\end{tabular*}
\end{table}

\subsection{The update-rate cap}\label{sec_supp_llm_cap}

Two independent mechanisms limit how many discrete state changes a layer may commit in one step, and both are applied per layer, per ternary component, per step. The first is a hard quota on deterministic flips. Given a cap $c$, at most $\lfloor c\,|\mathcal{W}| \rfloor$ coordinates may flip deterministically in a layer of $|\mathcal{W}|$ weights, and if more coordinates clear the adaptive threshold than the quota allows, only those with the largest $|\omega^n_{i}|$ are retained. The second is a ceiling on the stochastic flip probability, applied after the deterministic set has been resolved, which clamps $\exp(-|\omega^n_{i}|/(\bar{G}^n T^n))$ from above at $c$. Because $W^{\mathrm{on}}$ and $W^{\mathrm{off}}$ are resolved by two separate calls with independent random draws, the effective per-layer budget on changes to the effective ternary weight is up to $2c|\mathcal{W}|$, which is why cumulative flip counts can exceed the parameter count in uncapped runs.

In all experiments reported here the two mechanisms were set to the same value and varied together, so their individual contributions to the scale-dependent behaviour of Section~\ref{sec_supp_cap} are not separated. A cap below $1/|\mathcal{W}|$ truncates the deterministic quota to zero and disables deterministic flips for that layer entirely.

\section{Update-rate cap sweep}\label{sec_supp_cap}

The update-rate cap is the one hyperparameter whose optimal value we find does not transfer across model scale or training regime. This section reports the full sweep. All runs are $2{,}000$ steps with the configuration of Section~\ref{sec_supp_llm} and differ only in the cap.

\subsection{From-scratch pre-training}\label{sec_supp_cap_scratch}

\begin{table}[h]
\centering
\caption{Update-rate cap sweep for from-scratch pre-training. Tail-100 is the mean training loss over the final $100$ steps. Best tail-100 per size in bold.}\label{tab_supp_cap}
\footnotesize
\begin{tabular*}{\textwidth}{@{\extracolsep{\fill}}llccc}
\toprule
Model & Cap & Mean loss & Tail-100  \\
\midrule
MMFree-370M & uncapped   & $7.2055$ & $6.7011$  \\
MMFree-370M & $10^{-1}$  & $6.7278$ & $\mathbf{6.1271}$  \\
MMFree-370M & $10^{-2}$  & $6.6890$ & $6.1819$  \\
MMFree-370M & $10^{-3}$  & $6.8312$ & $6.4544$  \\
\midrule
MMFree-1.3B & uncapped   & $7.0237$ & $6.5612$  \\
MMFree-1.3B & $10^{-1}$  & $6.7439$ & $6.1921$  \\
MMFree-1.3B & $10^{-3}$  & $6.3630$ & $\mathbf{5.9684}$ \\
\midrule
MMFree-2.7B & $10^{-1}$  & $7.0736$ & $6.6201$  \\
MMFree-2.7B & $10^{-3}$  & $6.4871$ & $\mathbf{5.9846}$  \\
\bottomrule
\end{tabular*}
\end{table}

Three observations follow from Table~\ref{tab_supp_cap}. First, capping is necessary. At both sizes where an uncapped run exists it is the worst configuration on loss while also committing the most discrete updates, so there is no tradeoff to argue at that end of the range. Second, the optimum inverts with scale. At $370$M the best setting is $10^{-1}$, better than $10^{-3}$ by $0.327$ nats, whereas at $1.3$B the ordering reverses and $10^{-3}$ is better than $10^{-1}$ by $0.223$ nats, and at $2.7$B by $0.636$ nats. The preference for a tight budget therefore strengthens with model size, and no single default transfers. Third, at the two larger sizes the tight cap dominates outright, winning on loss while committing roughly $24\times$ fewer discrete updates than the loose setting. The quality-versus-traffic tradeoff that is visible at $370$M does not exist at the scales where the memory-traffic argument of Section~\ref{sec_res_hw} is most consequential.

\subsection{Continued pre-training}\label{sec_supp_cap_cpt}

The cap behaves differently when training resumes from a converged checkpoint. Table~\ref{tab_supp_cpt} reports every arm. Under the continued pre-training protocol of Section~\ref{sec_res_scaling}, the setting that is best for from-scratch $370$M pre-training, $10^{-1}$, is catastrophic, diverging to an evaluation loss of $7.1105$ against an untouched-checkpoint value of $3.1721$ and committing $1.54\times10^{8}$ flips over $100$ steps. The $10^{-3}$ arm shows the same mechanism in milder form, beginning far above the other arms and spending the run repairing early damage rather than improving monotonically. 
\begin{table}[h]
\centering
\caption{Continued pre-training from a common quantized checkpoint, $100$ steps, held-out evaluation loss. All arms share the same data, schedule, and continuous-parameter optimizer, and differ only in the ternary weight update rule. The initial loss difference is attributed to initialization of the scaling factors for both arms.}\label{tab_supp_cpt}
\footnotesize
\begin{tabular*}{\textwidth}{@{\extracolsep{\fill}}lccr}
\toprule
Update rule & Eval loss at step 0 & Final eval loss & $\Delta$ vs STE \\
\midrule
BitNet STE                                  & $3.1721$ & $3.1644$ & --- \\
$\Sigma\Delta$-Sampler, cap $10^{-5}$            & $3.1951$ & $3.0800$ & $-0.084$ \\
$\Sigma\Delta$-Sampler, cap $10^{-4}$            & $3.1950$ & $\mathbf{3.0241}$ & $\mathbf{-0.140}$ \\
$\Sigma\Delta$-Sampler, cap $10^{-3}$            & $3.1951$ & $3.0913$ & $-0.073$ \\
$\Sigma\Delta$-Sampler, cap $10^{-1}$            & $3.1950$ & $7.1105$ & $+3.946$ \\
\bottomrule
\end{tabular*}
\end{table}

The interpretation is that from-scratch training has no solution to preserve and benefits from aggressive discrete search, whereas continued pre-training begins at a good optimum that a high flip rate destroys faster than $100$ steps of gradient information can repair. The optimal cap therefore depends jointly on model scale and training regime.

\section{Training-state and memory-traffic accounting}\label{sec_supp_mem}

This section gives the derivations behind the two hardware results of Section~\ref{sec_res_hw}.

\subsection{Per-parameter training state}\label{sec_supp_mem_static}

The accounting covers persistent parameter state only, comprising the weight, the gradient, adaptive-optimizer moments, and per-algorithm auxiliary buffers. Intermediate activations are excluded because they are identical across optimizers under matched architecture and batch size, and because the activation path is shared between $\Sigma\Delta$-Sampler and the straight-through baseline in the language-model setting.

A full-precision or straight-through parameter carries four fp32 quantities, namely the weight, the gradient, and the AdamW first and second moments, for $16$ bytes. The straight-through case is identical to full precision here because the latent weight it maintains is itself a full-precision parameter held in the optimizer. A $\Sigma\Delta$-Sampler discrete parameter carries the ternary code as the pair $W^{\mathrm{on}}, W^{\mathrm{off}}$ packed at one byte per component, and the integrator $\omega^n$ in the existing gradient buffer at fp32, for $6$ bytes. The AdamW moments are structurally absent, since the annealing rule of Eq.~\eqref{eq_method_update_dynamics_full} consults only the current integrator state and a threshold, and the random draws consumed by the stochastic branch are transient within the anneal step and require no persistent allocation. Continuous parameters remain under AdamW at $16$ bytes.

Aggregating over each model's inventory gives the totals reported in Fig.~\ref{fig4}d. For a model with $N_d$ discrete and $N_c$ continuous parameters the ratio is
\begin{equation}
    \frac{M_{\Sigma\Delta}}{M_{\mathrm{FP}}} \;=\; \frac{6 N_d + 16 N_c}{16 (N_d + N_c)},
    \label{eq_supp_mem_ratio}
\end{equation}
which approaches $6/16 = 0.375$ as the discrete fraction approaches unity. The residual is carried entirely by the continuous parameters, which is why the saving grows with the discrete fraction and why it is larger on the language models, where $82.4\%$ to $93.9\%$ of parameters are ternary, than on ResNet-18, where the per-channel scaling parameters are proportionally more numerous.

\subsection{DRAM transaction model for weight updates}\label{sec_supp_mem_dram}

The traffic comparison is restricted to weight updates. The integrator is read and written densely at every step, but so is the gradient buffer of a full-precision or straight-through optimizer, and the two are the same size, so that component is common to both methods and cancels. What differs is the weight. A full-precision or straight-through optimizer applies a non-zero update to every weight at every step and therefore rewrites the entire weight tensor, whereas a $\Sigma\Delta$-Sampler discrete weight is written only when a flip is committed.

Various HBM(high bandwidth memory) technology provides configurable bandwidth for off-chip memory accesses. For this particular work, we assume they are fixed at a granularity of $32$ bytes. Writing a dense FP32 weight tensor of $N$ parameters therefore costs
\begin{equation}
    \Gamma_{\mathrm{FP}} \;=\; \frac{N}{S_{\mathrm{FP}}}, \qquad S_{\mathrm{FP}} = 8
    \label{eq_supp_dram_fp}
\end{equation}
transactions per step, with $S_{\mathrm{FP}}$ the number of weights per sector. For $\Sigma\Delta$-Sampler, writing $f(n)$ flipped one-byte codes costs
\begin{equation}
    \Gamma_{\Sigma\Delta}(n) \;=\; \min\!\left(f(n),\; \frac{N}{S_{\Sigma\Delta}}\right), \qquad S_{\Sigma\Delta} = 32,
    \label{eq_supp_dram_sd}
\end{equation}
where the first argument charges one full sector per flipped weight, granting no credit when two flipped weights share a sector, and the second saturates to the dense cost once flips exceed one per sector.

Two aspects of this construction are conservative against $\Sigma\Delta$-Sampler. Charging a full sector per flip is the worst case for a sparse and spatially uncorrelated update pattern, which is what the blue-noise stochasticity of Section~\ref{sec_supp_algo_bn} is designed to produce, so realized coalescing can only reduce $\Gamma_{\Sigma\Delta}$ below the reported value. Taking the baseline weight tensor as bf16 rather than the fp32 tensors the runs actually maintain halves the traffic attributed to it. The flip counts $f(n)$ are measured per step during the same runs whose loss trajectories appear in Fig.~\ref{fig4}b; only the conversion from flips to transactions is analytical. These figures are not hardware performance counters, and the model accounts for weight-update traffic alone, excluding activations, gradients, optimizer state, and collective communication.


\subsection{Per-third transaction rates}\label{sec_supp_mem_thirds}

Main-text Fig.~4d reports the mean transaction rate over each third of the run, normalized by training steps so that runs of different length remain comparable. The decrease from the early to the late third is the crystallization effect of Section~\ref{sec_res_benchmarks} expressed in traffic terms.



\subsection{Traffic as a function of the update-rate cap}\label{sec_supp_mem_cap}

Because the cap directly bounds the number of committed flips, it also bounds the weight-update traffic. Table~\ref{tab_supp_traffic_cap} reports both quantities for MMFree-370M against a full-precision reference of $19{,}267{,}584$ transactions per step. The traffic reduction spans a factor of forty across the sweep while the loss varies by $0.6$ nats, and the uncapped configuration is dominated on both axes.

\begin{table}[h]
\centering
\caption{Weight-update traffic and final loss against the update-rate cap for MMFree-370M. The full-precision reference is $19{,}267{,}584$ transactions per step.}\label{tab_supp_traffic_cap}
\footnotesize
\begin{tabular*}{\textwidth}{@{\extracolsep{\fill}}lrrc}
\toprule
Cap & Mean transactions/step & Reduction & Final loss \\
\midrule
uncapped   & $186{,}493$ & $103\times$   & $6.6555$ \\
$10^{-1}$  & $86{,}727$  & $222\times$   & $6.0654$ \\
$10^{-2}$  & $27{,}487$  & $701\times$   & $6.1407$ \\
$10^{-3}$  & $4{,}209$   & $4{,}578\times$ & $6.4093$ \\
\bottomrule
\end{tabular*}
\end{table}

\section{Hyperparameter sensitivity on CIFAR-10}\label{sec_supp_sweep}
 
This section reports a sweep over the $\Sigma\Delta$-Sampler hyperparameters on ternary VGG-8 / CIFAR-10, which is the setting we use to fix the defaults of Section~S2.3. Twenty-four configurations were screened for $100$ epochs and the top ten were rerun for $200$ epochs. Every run shares the same architecture, $\Sigma\Delta$ loop in \texttt{single\_omega} mode, hard dynamic threshold, gradient accumulation over two epochs, batch size $100$, step learning-rate decay every $70$ epochs, no dropout, and seed $42$, and differs from the baseline configuration only in the parameter named. Accuracy is CIFAR-10 validation accuracy in percent from a single seed per configuration.
 
\begin{table}[h]
\centering
\caption{$200$-epoch reruns of the ten configurations promoted from the screen, ordered by best validation accuracy. Each row differs from the baseline only in the parameter shown. Last-10 is the mean over the final ten epochs.}\label{tab_supp_sweep_e200}
\footnotesize
\begin{tabular*}{\textwidth}{@{\extracolsep{\fill}}lccc}
\toprule
Varied parameter & Best val & Final val & Last-10 mean \\
\midrule
\texttt{thld\_tau}$=5000$                        & $\mathbf{92.20}$ & $91.54$ & $91.76$ \\
\texttt{sd\_accum\_epochs}$=5$                   & $92.03$ & $91.74$ & $91.52$ \\
\texttt{temp\_decay}$=20$                        & $92.03$ & $91.86$ & $91.76$ \\
\texttt{thld\_target\_pruned\_init}$=0.3$        & $92.01$ & $91.68$ & $91.64$ \\
\texttt{wd}$=0$                                  & $91.99$ & $91.78$ & $91.70$ \\
\texttt{ema\_beta\_log}$=0.9$                    & $91.94$ & $91.31$ & $91.35$ \\
\texttt{thld\_gain}$=0.5$, \texttt{target\_pruned}$=0.9$ & $91.93$ & $91.43$ & $91.45$ \\
\texttt{thld\_gain}$=0.5$, \texttt{target\_pruned}$=0.5$ & $91.91$ & $91.84$ & $91.57$ \\
\texttt{thld\_gain}$=1.5$, \texttt{target\_pruned}$=0.5$ & $91.91$ & $91.60$ & $91.32$ \\
\texttt{sd\_accum\_epochs}$=20$                  & $91.86$ & $91.70$ & $91.61$ \\
\bottomrule
\end{tabular*}
\end{table}
 
\begin{table}[h]
\centering
\caption{$100$-epoch screen over all twenty-four configurations, grouped by the parameter varied. The row marked $\dagger$ is the baseline configuration. Rows marked $\ast$ were promoted to the $200$-epoch reruns of Table~\ref{tab_supp_sweep_e200}.}\label{tab_supp_sweep_smoke}
\footnotesize
\begin{tabular*}{\textwidth}{@{\extracolsep{\fill}}llcc}
\toprule
Group & Setting & Best val & Final val \\
\midrule
Accumulation window & \texttt{sd\_accum\_epochs}$=5$ $\ast$  & $91.24$ & $90.57$ \\
                    & \texttt{sd\_accum\_epochs}$=10$        & $91.18$ & $90.91$ \\
                    & \texttt{sd\_accum\_epochs}$=20$ $\ast$ & $91.36$ & $90.99$ \\
\midrule
Threshold gain      & $0.5$, target $0.5$ $\ast$             & $91.32$ & $91.14$ \\
$\times$ AGC target & $0.5$, target $0.7$                    & $91.07$ & $91.05$ \\
                    & $0.5$, target $0.9$ $\ast$             & $\mathbf{91.44}$ & $91.03$ \\
                    & $0.95$, target $0.5$                   & $91.05$ & $90.51$ \\
                    & $0.95$, target $0.7$ $\dagger$         & $91.19$ & $90.90$ \\
                    & $0.95$, target $0.9$                   & $91.17$ & $91.05$ \\
                    & $1.5$, target $0.5$ $\ast$             & $91.39$ & $90.91$ \\
                    & $1.5$, target $0.7$                    & $91.07$ & $90.82$ \\
                    & $1.5$, target $0.9$                    & $91.18$ & $90.87$ \\
\midrule
AGC target ramp     & \texttt{target\_pruned\_init}$=0.1$    & $91.19$ & $90.49$ \\
                    & \texttt{target\_pruned\_init}$=0.3$ $\ast$ & $91.26$ & $90.91$ \\
\midrule
Threshold time const.& \texttt{thld\_tau}$=1000$             & $91.03$ & $90.48$ \\
                    & \texttt{thld\_tau}$=5000$ $\ast$       & $91.29$ & $90.78$ \\
\midrule
Baseline EMA        & \texttt{ema\_beta\_log}$=0.5$          & $91.18$ & $90.63$ \\
                    & \texttt{ema\_beta\_log}$=0.9$ $\ast$   & $91.30$ & $91.00$ \\
\midrule
Temperature decay   & \texttt{temp\_decay}$=20$ $\ast$       & $91.23$ & $90.83$ \\
                    & \texttt{temp\_decay}$=50$              & $90.76$ & $90.48$ \\
\midrule
Continuous-parameter& \texttt{lr}$=0.005$                    & $90.95$ & $90.66$ \\
learning rate       & \texttt{lr}$=0.03$                     & $90.61$ & $90.49$ \\
\midrule
Weight decay        & \texttt{wd}$=0$ $\ast$                 & $91.27$ & $90.78$ \\
                    & \texttt{wd}$=5\times10^{-4}$           & $91.07$ & $91.04$ \\
\bottomrule
\end{tabular*}
\end{table}
 
The dominant observation is insensitivity. All twenty-four screened configurations fall within a band of $0.83$ percentage points, and the ten promoted configurations fall within $0.34$ percentage points after $200$ epochs, which is narrower than the roughly $0.5$ percentage point run-to-run spread we observe across seeds for well-behaved CIFAR-10 configurations. The nine-point grid over threshold gain and AGC pruned-fraction target spans only $0.39$ percentage points, which is the behaviour the automatic gain control is intended to produce, since the loop regulates the flip population toward its target regardless of where the gain is set. The one parameter that moves the result consistently is the learning rate of the continuous scaling parameters, where both off-baseline settings rank in the bottom three, so the residual sensitivity of the method sits in the continuous half of the optimizer rather than in the annealing schedule. The best configuration reaches $92.20\%$ with \texttt{thld\_tau}$=5000$, and we adopt the baseline of Section~S2.3 rather than this value because the difference is within seed noise.
 
Two caveats apply. Each configuration was run at a single seed, so the ordering within either table should be read as a shortlist rather than a ranking. The screen also under-rates the schedule parameters by construction, since \texttt{temp\_decay}, \texttt{thld\_tau} and the AGC target ramp are indexed by optimizer step, so a $100$-epoch run under-cools the temperature and under-ramps the target relative to the $200$-epoch runs. This is visible in the two parameters whose screen and rerun ranks disagree most, \texttt{thld\_tau}$=5000$ and \texttt{temp\_decay}$=20$, both of which rise to the top three once given the full schedule.


\section{GPU power and energy during training}\label{sec_supp_gpupower}

\begin{figure}[h!]
\centering
\includegraphics[width=0.95\textwidth]{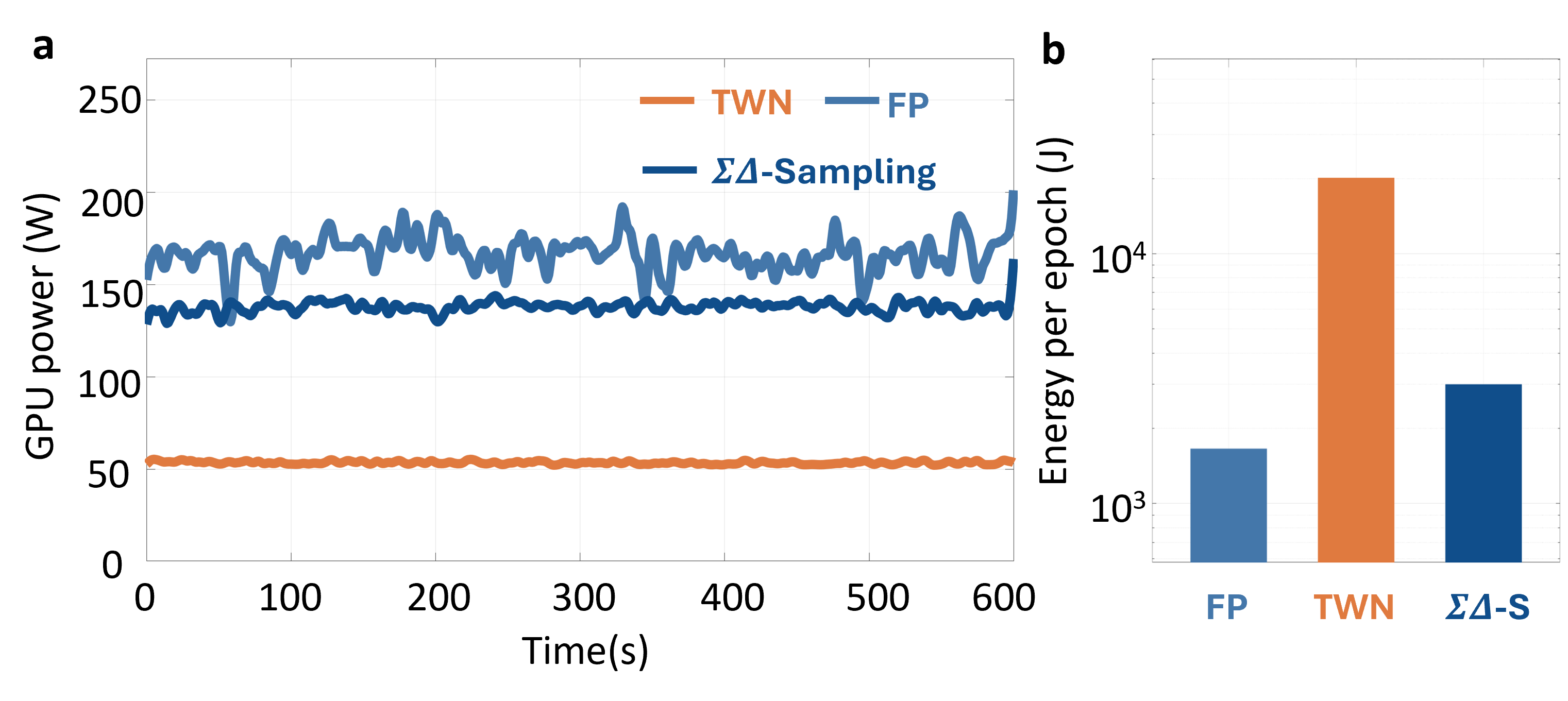}
\caption{\textbf{GPU power benchmarking of three optimizers on VGG-7 / CIFAR-10.} (a) Instantaneous board power against wall-clock time. Full precision and $\Sigma\Delta$-Sampler draw comparable power at comparable device utilization, whereas TWN sustains only $3.2\%$ of the above-idle power and $2.6\%$ of the step throughput of the full-precision baseline. (b) Energy per epoch. TWN is an order of magnitude above the other two, while $\Sigma\Delta$-Sampler incurs a $1.8\times$ overhead relative to full precision that we attribute to unfused kernels rather than to the update rule.}\label{sfig3}
\end{figure}

Power was sampled on a single A100 with $40$ GB of device memory under matched model, dataset, and batch size, with each optimizer run for $600$ seconds of wall-clock training. Energy per epoch is computed from the measured mean board power and the measured step rate at $500$ steps per epoch. Idle board power is $49.8$ W and is reported separately because it is common to all three arms and is not attributable to the optimizer.

\begin{table}[h]
\centering
\caption{Measured power, throughput, and derived energy for the three optimizers. Above-idle power subtracts the $49.8$ W idle draw of the device. Energy figures use total board power.}\label{tab_supp_power}
\footnotesize
\begin{tabular*}{\textwidth}{@{\extracolsep{\fill}}lrrrrrrr}
\toprule
Optimizer & Mean power & Above idle & Util. & Steps / $600$ s & s/epoch & J/epoch & J/step \\
\midrule
Full precision      & $166.6$ W & $116.8$ W & $42.3\%$ & $30{,}133$ & $9.96$  & $1{,}659$  & $3.32$ \\
TWN                 & $53.5$ W  & $3.7$ W   & $34.3\%$ & $797$      & $376.5$ & $20{,}145$ & $40.29$ \\
$\Sigma\Delta$-Sampler   & $137.9$ W & $88.1$ W  & $52.7\%$ & $13{,}776$ & $21.78$ & $3{,}004$  & $6.01$ \\
\bottomrule
\end{tabular*}
\end{table}

The quantity this measurement speaks to is implementation efficiency rather than algorithmic cost, since all three arms execute the same architecture on the same data and differ only in how the weight update is realized in kernels. Read that way, $\Sigma\Delta$-Sampler is on par with the full-precision baseline. It draws $88.1$ W above idle against $116.8$ W and sustains the highest device utilization of the three at $52.7\%$, indicating that the annealing update keeps the device busy with work rather than stalling it. TWN sits at the opposite end, drawing only $3.7$ W above idle while still registering $34.3\%$ utilization, which illustrates that utilization measures the fraction of time some kernel is resident rather than the arithmetic throughput achieved, and a pipeline that is latency-bound on many small launches can appear occupied while performing little work.

Energy per epoch does not follow power, because it is dominated by time to completion. TWN requires $376.5$ seconds per epoch and consequently spends $12.1\times$ the energy of the full-precision baseline despite drawing the least power of the three. $\Sigma\Delta$-Sampler completes an epoch in $21.78$ seconds against $9.96$ seconds for full precision, and the resulting $1.8\times$ energy overhead is entirely a throughput deficit rather than a power deficit. We attribute that deficit to the present implementation rather than to the update rule. The annealing step currently materializes the integrator, the threshold comparison, and the flip mask as separate tensors and launches them as separate kernels, so each discrete update traverses device memory several times where a fused kernel would traverse it once. The arithmetic the update rule performs per parameter is a comparison against a threshold and a conditional sign change, which is strictly cheaper than the multiply-accumulate and two moment updates of an adaptive full-precision optimizer, so we expect a fused implementation to be at least competitive with the full-precision baseline on time and therefore on energy. 

\section{Sensitivity to residual connectivity}\label{sec_supp_resconnect}

\begin{figure}[h!]
\centering
\includegraphics[width=0.95\textwidth]{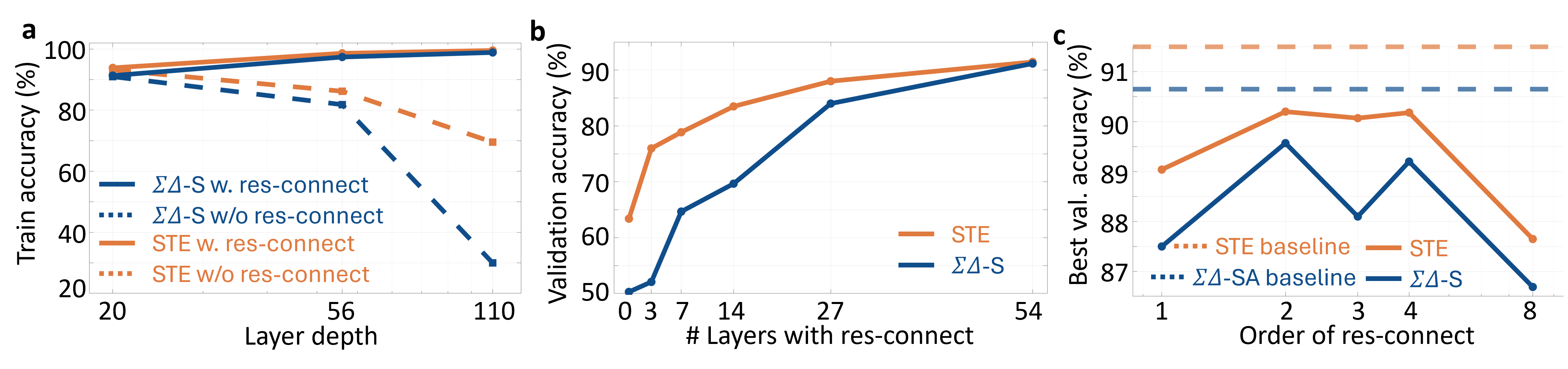}
\caption{\textbf{Sensitivity of optimizers to residual connectivity on CIFAR-10.} (a) Final training accuracy of STE and $\Sigma\Delta$-Sampler on CIFAR-style ResNets of increasing depth, with and without residual connections. (b) Peak validation accuracy of ResNet-56 against the number of blocks retaining a skip, counted from the input. (c) Test accuracy of the same architecture under higher-order residual connections.}\label{sfig2}
\end{figure}

All runs use CIFAR-style ResNets of depth $6n{+}2$ with ternary weights throughout, trained for $200$ epochs at batch size $128$ and learning rate $0.01$ decayed by $0.1$ every $50$ epochs at seed $42$. The two arms differ only in the discrete update rule, with the STE baseline maintaining a latent fp32 weight updated by AdamW through a clipped straight-through estimator with absmean ternarization, and $\Sigma\Delta$-Sampler annealing the $W_{\mathrm{on}}/W_{\mathrm{off}}$ pair directly. Shape-changing skips use the parameter-free option-A shortcut rather than a $1\times1$ projection, so residual and plain networks are identical in parameter count, $3{,}638{,}852$ at depth $110$, and adding or removing skips changes only the gradient path. Panel~(b) retains a skip in the first $K$ blocks counted from the input, the end of the network farthest from the loss, while panel~(c) generalizes the reach of every skip to a residual order $R$,
\begin{equation}
x_i \;=\; \mathrm{block}_i(x_{i-1}) \;+\; \textstyle\sum_{r=1}^{R}\,
\mathrm{LN}\!\left(\mathcal{A}(x_{i-r})\right),
\label{eq_supp_resorder}
\end{equation}
so that each block sums the $R$ preceding block outputs rather than its immediate predecessor alone, with $R=1$ recovering the standard ResNet. The adapter $\mathcal{A}$ reshapes an earlier activation by the same option-A recipe, and the LayerNorm carries no affine transformation, so the parameter count is invariant in $R$. That normalization is necessary rather than cosmetic, since under a plain sum the residual stream follows a Fibonacci recursion whose geometric growth drives the depth-$110$ midpoint activation to $3.5\times10^{15}$ at $R{=}4$ on initialization alone.

Panel~(a) establishes the precondition. Removing every skip is almost free at depth $20$, below the depth at which skips matter, but the penalty grows sharply thereafter and grows faster for $\Sigma\Delta$-Sampler, from $12.4$~pp against $15.6$~pp at depth $56$ to $30.0$~pp against $68.9$~pp at depth $110$, where the discrete optimizer no longer fits the training set. Because this degradation is an optimization failure rather than a generalization one, panel~(a) reports training accuracy. Panel~(b) shows that the optimizers also differ in where skips must sit. Restoring skips to three of $54$ blocks nearest the input recovers $45\%$ of the plain-to-residual gap under STE, well beyond the $6\%$ of blocks involved, after which recovery saturates. The same three blocks return $4\%$ to $\Sigma\Delta$-Sampler, which catches up only once roughly half the network carries skips. A few input-side skips therefore suffice for straight-through training, whereas $\Sigma\Delta$-Sampler needs them distributed throughout.

Panel~(c) tests whether additional residual structure converts this dependence into an advantage, and it does not. Higher-order skips never exceed the standard ResNet for either optimizer. LayerNorm alone at $R{=}1$ costs $2.5$~pp for STE and $3.2$~pp for $\Sigma\Delta$-Sampler, consistent with a normalized skip no longer acting as an identity mapping, and raising the order recovers about half of that loss by $R{=}2$ and nothing beyond. Notably, the settings with the flattest gradient norm across depth at initialization are not the best trained, so improved gradient flow at initialization does not by itself predict trained accuracy. Across all three panels $\Sigma\Delta$-Sampler depends on residual structure at least as much as STE does, and adding structure does not reverse that ordering.

All cells are single-seed. Repeated configurations spread by $6$ to $15$~pp in the strongly degraded regime but only about $0.5$~pp with skips throughout, so the large effects above sit well outside run-to-run noise while differences below roughly $1$~pp between neighbouring values of $R$ do not. $\Sigma\Delta$-Sampler is additionally unstable at intermediate $K$, peaking at $64.7\%$ near epoch $95$ before falling to $35.6\%$ by epoch $200$ at $K{=}7$, whereas STE never declines by more than $1.1$~pp, which is why panel~(b) reports peak validation accuracy.

\newpage
\bibliography{sn-bibliography}
\end{document}